\documentclass{article}

\usepackage{arxiv}
\usepackage[T1]{fontenc}
\usepackage[utf8]{inputenc}
\usepackage{textcomp}
\usepackage{amsmath,amssymb,mathtools}
\usepackage{bm}
\usepackage{xcolor}
\usepackage{longtable,booktabs,array}
\usepackage{calc}
\usepackage{etoolbox}
\usepackage{graphicx}
\usepackage{microtype}
\usepackage[font=small,labelfont=bf,labelsep=period]{caption}
\usepackage[section]{placeins}
\usepackage{url}
\usepackage[authoryear,round]{natbib}
\usepackage{hyperref}
\definecolor{ReferenceRose}{HTML}{AD3154}
\hypersetup{colorlinks=true,citecolor=ReferenceRose,linkcolor=ReferenceRose,
  urlcolor=ReferenceRose,filecolor=ReferenceRose,pdfborder={0 0 0}}
\newcounter{displayanchor}
\AddToHook{env/equation/before}{\stepcounter{displayanchor}}

\DeclareRobustCommand{\Figref}[2][]{\hyperref[fig:#2]{\textbf{Fig.~\ref*{fig:#2}#1}}}
\DeclareRobustCommand{\Tabref}[1]{\hyperlink{tab:#1}{\textbf{Table~#1}}}
\DeclareRobustCommand{\Suppref}[2]{\textcolor{ReferenceRose}{\textbf{#2}}}

\makeatletter
\patchcmd\longtable{\par}{\if@noskipsec\mbox{}\fi\par}{}{}
\makeatother

\makeatletter
\def\maxwidth{\ifdim\Gin@nat@width>\linewidth\linewidth\else\Gin@nat@width\fi}
\def\maxheight{\ifdim\Gin@nat@height>\textheight\textheight\else\Gin@nat@height\fi}
\makeatother
\setkeys{Gin}{keepaspectratio}
\renewcommand{\undertitle}{A PREPRINT}
\renewcommand{\headeright}{A PREPRINT}
\renewcommand{\shorttitle}{Composition-Dependent Models for Digital Materials}

\title{Data-Driven Discovery of Composition-Dependent Constitutive Models for Hyperelasticity and Viscoelasticity of Digital Materials}

\author{
\begin{minipage}[t]{0.29\textwidth}
\centering\normalfont\small
\textbf{Josué García-Ávila}\\
Department of Mechanical Engineering\\
Columbia University\\
New York City, NY 10027, USA\\
\href{mailto:jg5020@columbia.edu}{\texttt{jg5020@columbia.edu}}
\end{minipage}
\And
\begin{minipage}[t]{0.29\textwidth}
\centering\normalfont\small
\textbf{Beijun Shen}\\
Department of Mechanical Engineering\\
Columbia University\\
New York City, NY 10027, USA\\
\href{mailto:bs3660@columbia.edu}{\texttt{bs3660@columbia.edu}}
\end{minipage}
\And
\begin{minipage}[t]{0.29\textwidth}
\centering\normalfont\small
\textbf{Manuel K. Rausch}\\
Departments of Aerospace Engineering and Engineering Mechanics, Biomedical Engineering, and Mechanical Engineering\\
University of Texas at Austin\\
Austin, TX 78712, USA\\
\href{mailto:manuel.rausch@utexas.edu}{\texttt{manuel.rausch@utexas.edu}}
\end{minipage}
\AND
\begin{minipage}[t]{0.29\textwidth}
\centering\normalfont\small
\textbf{Mary C. Boyce}\\
Department of Mechanical Engineering\\
Columbia University\\
New York City, NY 10027, USA\\
\href{mailto:mb3814@columbia.edu}{\texttt{mb3814@columbia.edu}}
\end{minipage}
\And
\begin{minipage}[t]{0.29\textwidth}
\centering\normalfont\small
\textbf{Adrián Buganza-Tepole\textsuperscript{*}}\\
Department of Mechanical Engineering\\
Columbia University\\
New York City, NY 10027, USA\\
\href{mailto:ab6035@columbia.edu}{\texttt{ab6035@columbia.edu}}
\\\textsuperscript{*}Corresponding author
\end{minipage}
}
\hypersetup{pdfauthor={Josué García-Ávila, Beijun Shen, Manuel K. Rausch, Mary C. Boyce, Adrián Buganza-Tepole},pdftitle={Data-Driven Discovery of Composition-Dependent Constitutive Models for Hyperelasticity and Viscoelasticity of Digital Materials}}
\date{}
\begin{document}
\maketitle

\begin{abstract}
Digital materials fabricated by multi-material 3D printing are designed as controlled mixtures of stiff and compliant constituents, yielding effective responses that span more than an order of magnitude in apparent stiffness and exhibit strongly nonlinear, composition-dependent, and rate-dependent dissipative behavior. Classical finite-strain viscoelastic models represent such behavior with closed-form strain energy functions for equilibrium and non-equilibrium stresses as well as evolution of internal variables, which may limit flexibility when a single constitutive model is expected to generalize across materials and loading rates. Here, we present a data-driven multi-material constitutive modeling framework that generalizes a formulation by Bergström and Boyce. The proposed framework retains the structure of the classical model, namely multiplicative kinematics, invariant-based strain-energy functions, and a scalar dissipative evolution law directed along the normalized nonequilibrium deviatoric stress. For the equilibrium branch, the data-driven discovery framework either directly predicts closed-form model parameters as functions of composition or automatically constructs a polyconvex strain-energy function using neural ordinary differential equations (NODEs). The nonequilibrium branch kinetics are learned similarly, either by directly identifying closed-form parameters across compositions or by using appropriately constrained artificial neural networks. Using multi-rate uniaxial compression data across multiple material compositions, we show that the proposed formulation captures rate-dependent stiffness and hysteresis across compositions while preserving thermodynamic consistency.
\end{abstract}


\section*{1. Introduction}

Constitutive modeling of nonlinear viscoelastic solids under large finite deformations remains a challenging, persistent problem in continuum mechanics, driven by continuous innovation in soft material design and novel fabrication techniques. An example is voxel-based multi-material composites fabricated via material jetting, where rigid and soft photopolymers are combined at the voxel scale to generate so-called digital materials \citep{Slesarenko2018,Yuan2021}. Micro- or macro-scale voxelization and stochastic distribution between layers allow the creation of materials with an effective mechanical response distinct from that of their base constituents (see \Figref[A]{1}), enabling a wide range of behaviors through the sole variation of composition and spatial distribution. This functionalization can be extended to applications requiring specific stress redistribution with tunable soft-rigid transition interfaces through continuous blending of polymeric components \citep{Saldvar2023,Roh2024}. These libraries of base materials thus motivate the development of new constitutive frameworks that describe mechanical behavior continuously across related compositions, rather than treating each mixture as an isolated, independent material.

Despite decades of development, formulating models capable of consistently describing the coupling between geometric nonlinearity, the time dependence of internal variables, and their associated energy dissipation across a wide range of materials and adhering to basic mathematical and physical constraints remains a challenge. Although many constitutive models exist for nonlinear elastic and inelastic materials, model selection and robust parameter identification for phenomenological models remain difficult inverse problems \citep{Xiang2020,Dal2021}. In this context, models inspired by statistical mechanics are especially relevant. By imposing explicit mechanisms associated with molecular networks, these formulations restrict the space of admissible responses and, in doing so, reduce non-uniqueness and improve calibration. They can also be formulated to satisfy requirements such as objectivity and, in specific cases such as the Arruda-Boyce eight-chain model, polyconvexity under appropriate parameter restrictions \citep{Hartmann2003}. At the same time, these advantages depend on the adequacy of the assumed mechanisms. Overly restrictive assumptions can limit accuracy under nonlinear, rate-dependent, or history-dependent loading \citep{Dornheim2023}.

The micromechanics-based Arruda-Boyce model describes the hyperelastic response of polymer networks as a function of the chain stretch \(\lambda_{chain}\) (obtained by assuming affine deformation of the polymer chains with the principal stretches) using only two parameters \citep{Arruda1993}. There are extensions of this model, such as the Flory-Erman model, which introduced an energy correction associated with topological constraints and non-affine fluctuations \citep{Flory1982}. Hybrid formulations of these two models have also been proposed to improve predictions in multiaxial states, particularly in equibiaxial extension \citep{Boyce2000}. For finite viscoelasticity, the Bergström-Boyce model \citep{Bergstrm1998} couples a hyperelastic equilibrium network with a non-equilibrium network. The equilibrium branch stores energy as a function of the Cauchy-Green tensor \(\mathbf{C}\) or its invariants. The non-equilibrium material response depends on history and, therefore, cannot be reconstructed from an instantaneous measure of deformation such as \(\mathbf{C}\). General thermodynamically consistent formulations introduce a set of internal variables \(q\) that act as carriers of memory \citep{Coleman1967}:

\begin{equation}
\psi = \widehat{\psi}\left( \mathbf{C,}q \right).
\tag{1.1}\label{eq:1.1}
\end{equation}

This general structure does not uniquely prescribe the choice, physical meaning, or evolution of the internal variables. However, in a particularly influential class of finite viscoelasticity models \citep{Reese1998a}, the canonical internal state is represented by viscous strains, \({q\mathbf{\equiv}\mathbf{F}}_{v}\), via the multiplicative split of the deformation gradient, see also \citep{Eckart1948,Kondo1950,Bilby1958,Krner1959,Lee1969} (see \Figref[B]{1}):

\begin{equation}
\mathbf{F} = \mathbf{F}_{e}\text{\,}\mathbf{F}_{v}.
\tag{1.2}\label{eq:1.2}
\end{equation}

Note that \(\mathbf{F}\) is a measurable observable quantity, while \(\mathbf{F}_{e}\) and \(\mathbf{F}_{v}\) are internal state tensors that cannot be measured directly.

The strain energy density function \(\psi\) can then be interpreted as two networks acting in parallel (see \Figref[C]{1}): an equilibrium (\(\psi^{EQ}\)) response and a time-dependent nonequilibrium response (\(\psi^{NEQ}\)). For many elastomers, the energy is assumed to be purely isochoric. In that case, after imposing the volumetric constraint \(J = \det\mathbf{F} = 1\), the free energy can be expressed as a sum of the two contributions

\begin{equation}
\psi = \psi_{iso}^{EQ}\left( \mathbf{C} \right) + \psi_{iso}^{NEQ}\left( \mathbf{C}_{e} \right),
\tag{1.3}\label{eq:1.3}
\end{equation}

with \(\mathbf{C}_{e} = \mathbf{F}_{e}^{T}\mathbf{F}_{e}\). To close the system of equations, a constitutive evolution of the internal variables \(\mathbf{F}_{v}\) is needed, which must satisfy the second law of thermodynamics. Locally and isothermally, this is expressed by the condition of non-negative internal dissipation in the Clausius-Duhem inequality:

\begin{equation}
\mathcal{D}_{int} \geq 0.
\tag{1.4}\label{eq:1.4}
\end{equation}

A suitable family of analytical models that satisfy the Clausius-Duhem inequality as well as objectivity are based on connecting the rate of change of the viscous strains to an effective measure of the non-equilibrium stress via power laws \citep{Bergstrm1998,Reese1998b,Haupt2002}. However, in practice, several non-equilibrium branches and addition of other phenomena such as softening or hardening lead to very nonlinear behavior, a relatively large number of parameters, and challenges in parameter identification. The problem is compounded when the same modeling framework is expected to work across a wide family of materials such as voxel-based composites. Therefore, despite the rigor of the thermodynamic framework, the remaining challenges are determining the constitutive form of \(\psi\) and specifying the kinetics of the internal variables.

In recent years, data-driven frameworks, particularly artificial neural networks (NNs), have emerged with the promise of approximating complex constitutive relations directly from experimental data. However, many proposed neural constitutive models lack the structural guarantees required by continuum mechanics: unconstrained formulations may violate increasing energy monotonicity, objectivity, material symmetries, and, critically, the second law of thermodynamics \citep{Linden2023}. The direct prediction of stress using NNs without an explicit separation between energy storage and dissipation leads to black-box models that are difficult to interpret and unreliable outside the training domain.

Based on this consideration, various physically constrained neural frameworks have been proposed to learn constitutive functions. These frameworks, often referred to as part of a big family called physics-informed or physics-augmented neural networks (PINNs or PANNs respectively), incorporate physical principles or constraints into neural networks \citep{Karniadakis2021,Fuhg2024}. For example, constitutive artificial neural networks (CANNs) learn non-negative coefficients on convex bases \citep{Linka2021}; input convex neural networks (ICNNs) enforce convexity through non-negative weights and composition of convex monotonic activation functions \citep{Fuhg2022,Klein2022a}. In contrast to these approaches, neural ordinary differential equations (NODEs) avoid interpolating the energy itself and instead interpolate the derivative functions directly, which allow the assembly of the stress tensor without additional differentiation. The NODE approach satisfies convexity conditions \emph{a priori} through monotonicity and non-negativity of the strain energy derivatives \emph{via} the NODE architecture. \citep{Tac2022,BuganzaTepole2025}.

While PANNs have been widely developed for hyperelasticity, the development of physics-augmented data-driven methods in viscoelasticity has been slower. This is largely due to two major challenges: (i) ensuring non-negative energy dissipation by design and (ii) managing the high computational cost of parameter optimization \citep{Ta2023}. Recent work has begun to address these limitations. For digital materials, Yang et al. combined a partially input convex neural network (pICNN) to learn the composition-dependent hyperelastic strain energy function paired with a quasi-linear viscoelastic (QLV) formulation \citep{Yang2025}. Although composition-aware, the time-dependent response remains embedded in a prescribed QLV structure rather than in explicitly evolving finite-strain internal variables. More recently, Jones and Fuhg proposed an internal state variable framework in which hidden internal states are inferred directly from data rather than prescribed \emph{a priori}. In their work, both stress and internal state evolution are learned under thermodynamic and convexity constraints \citep{Jones2025}. However, the inferred states remain latent and difficult to interpret, and the framework was not developed for a continuous family of material compositions. These limitations leave open the need for a robust multi-material framework combining an interpretable structure, flexible neural constitutive functions, and automatic satisfaction of thermodynamic admissibility.

In this work, we propose a unified constitutive framework for finite nonlinear hyperelasticity and viscoelasticity that reconciles classical analytical models with the flexibility of data-driven techniques to capture the behavior of a large class of voxel-based composites. We preserve the thermodynamic structure of large deformation viscoelasticity by relying on multiplicative kinematics of the deformation gradient and a viscoelastic flow proportional to the normalized non-equilibrium stress. The multi-material framework leverages NNs to either learn composition-dependent parameters of analytical strain energies and flow rules, or describe the constitutive models directly through NODEs and positive NNs conditioned on material features (see \Figref[D]{1}). Ultimately, the multi-material framework can capture how the rate-dependent response varies systematically with composition across apparent stiffness spanning more than one order of magnitude and nearly three orders of magnitude in strain rate (see \Figref[E]{1}).

\begin{figure}[htbp]
\centering
\includegraphics[width=\linewidth,keepaspectratio]{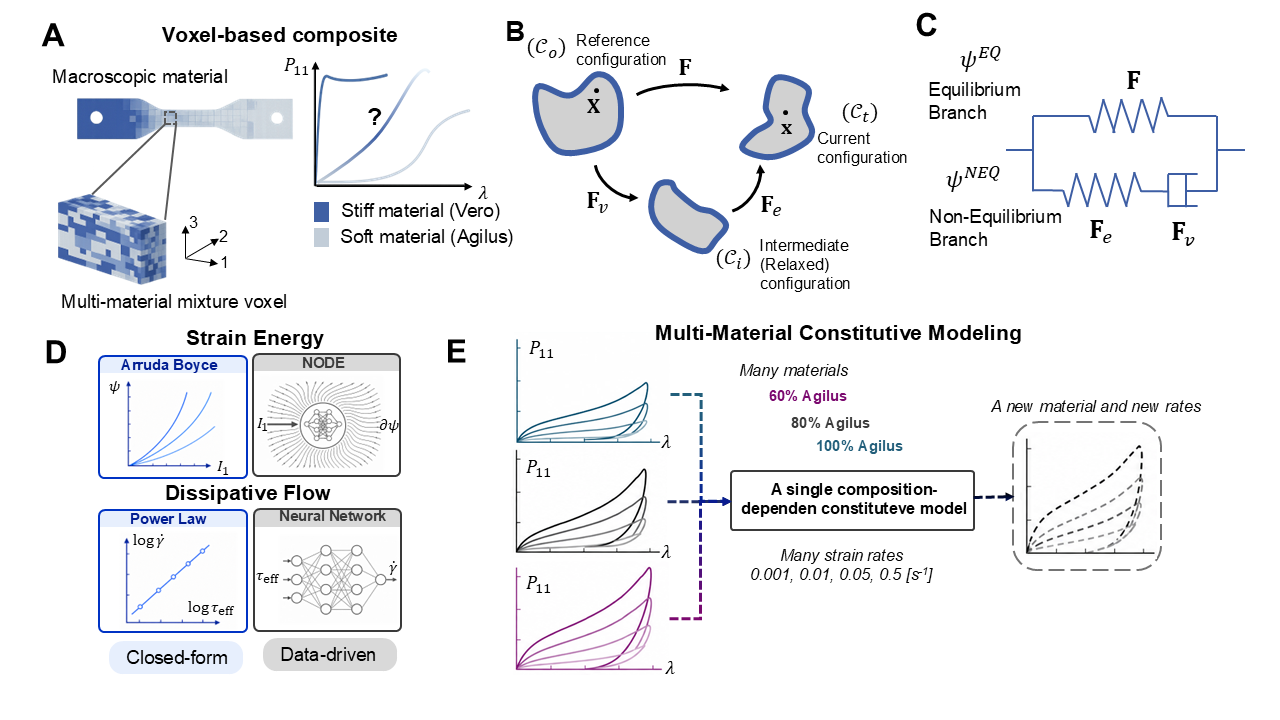}
\caption{\textbf{(A)} Voxel-based multi-material design produces macroscopic responses that differ from those of the base constituents. \textbf{(B)} A body undergoing a motion from the reference configuration \(\mathcal{C}_{0}\) to the current configuration \(\mathcal{C}_{t}\) through an intermediate configuration \(\mathcal{C}_{\mathcal{i}}\). \textbf{(C)} Classical finite-viscoelasticity is interpreted through parallel equilibrium and one or more non-equilibrium network representations. \textbf{(D)} In the proposed constitutive framework, deformation data \(\mathbf{F}\) and material descriptors \(Z\) drive the stress and evolution of internal state variables via two scalar constitutive laws: strain energy and a dissipative scalar flow \(\dot{\gamma}\). The energy is represented either by Arruda--Boyce or by NODE, whereas the evolution law is represented either by a power law or by a positive neural network. (\textbf{E).} A shared multi-material constitutive law conditioned on \(Z\) is trained across materials and strain rates, enabling interpolation to unseen compositions and rates. Note that the representative responses differ not only in magnitude, but also in initial slope, residual strain, and rollover stress.}
\label{fig:1}
\end{figure}

\section*{2. Constitutive framework for finite-strain viscoelasticity}

\subsection*{2.1 Branch-wise multiplicative kinematics}

To describe the rate-dependent behavior under finite strain, each non-equilibrium branch is assigned its own internal viscous deformation through multiplicative decomposition of the deformation gradient \(\mathbf{F}\mathbf{=}\frac{\partial\mathbf{x}}{\partial\mathbf{X}}\), where \(\mathbf{x}\mathcal{= M}\left( \mathbf{X},t \right)\) is the deformation map between the reference and current configuration \(\mathcal{C}_{t}\). For each rheological branch \emph{r} we write

\begin{equation}
\mathbf{F} = \mathbf{F}_{e}^{(r)}\mathbf{F}_{v}^{(r)},\ \ r \in \mathcal{R}_{neq},
\tag{2.1}\label{eq:2.1}
\end{equation}

where \(\mathbf{F}_{e}^{(r)}\) is the branch elastic distortion, and \(\mathbf{F}_{v}^{(r)}\) is the internal viscous distortion state (\(q \equiv \mathbf{F}_{v}^{(r)}\)) carrying the branch memory. In the present work, the nonequilibrium branch set is \(\mathcal{R}_{neq} = \left\{ B,C \right\}\), while branch \(A\) denotes the equilibrium branch. The elastic part of each branch is captured by the right and left Cauchy-Green tensors,

\begin{equation}
\mathbf{C}_{e}^{(r)} = \mathbf{F}_{e}^{(r)T}\mathbf{F}_{e}^{(r)}\quad\text{and}\quad\mathbf{B}_{e}^{(r)} = \mathbf{F}_{e}^{(r)}\mathbf{F}_{e}^{(r)T}.
\tag{2.2}\label{eq:2.2}
\end{equation}

The multiplicative structure in \textbf{Eq. 2.1} provides a geometrically meaningful representation of internal history. Additional details on the kinematics and the reduction to uniaxial incompressible deformation are collected in \textbf{Appendix A}.

\subsection*{2.2. Free-energy decomposition and multi-material dependence}

The total Helmholtz free energy is decomposed into an equilibrium contribution from branch \(A\) and a sum of nonequilibrium contributions,

\begin{equation}
\psi = \psi^{(A)}\left( \mathbf{C};Z \right) + \sum_{r \in \mathcal{R}_{neq}}^{}\psi^{(r)}\left( \mathbf{B}_{e}^{(r)};Z \right),
\tag{2.3}\label{eq:2.3}
\end{equation}

where \(Z\) is a conditioning vector that does not depend on deformation but encodes the material descriptors defining each sample in the multi-material setting, for example, hardness \((H)\), composition \((P)\), photopolymer family 1 \(\left( d_{1} \right)\), and photopolymer family 2 \({(d}_{2})\),

\begin{equation}
Z = \lbrack H,\,P,\,d_{1},\,d_{2}\rbrack.
\tag{2.4}\label{eq:2.4}
\end{equation}

In this work, all materials share the same base-material recipe (\(d_{1},d_{2}\)). Because \(H\) and \(P\) are correlated, we use only \(Z = \lbrack P\rbrack\), conditioning the model on the composition of one base material. Nevertheless, \(Z\) can be extended to include additional relevant material descriptors.

The first Piola--Kirchhoff stress follows from the free-energy derivative with respect to \(\mathbf{F}\)

\begin{equation}
\bm{P} = \frac{\partial\psi}{\partial\mathbf{F}} - p^{*}\mathbf{F}^{\mathbf{-}\mathbf{T}} = \bm{P}_{eq}^{(A)} + \sum_{r \in \mathcal{R}_{neq}}^{}\bm{P}_{neq}^{(r)} - p^{*}\mathbf{F}^{- T},
\tag{2.5}\label{eq:2.5}
\end{equation}

where \(p^{*}\) denotes the Lagrange multiplier enforcing the incompressibility constraint \(J = 1\). The additive stress decomposition under incompressibility follows from the free-energy split and is detailed further in \textbf{Appendix B.2} and \textbf{Appendix C.2.} Note that, given the first Piola Kirchoff stress tensor, standard push-forward or pull-back operations with \(\mathbf{F}\) can be used to define the common stress measures: second Piola Kirchhoff stress \(\mathbf{S}\), Cauchy stress \(\mathbf{\sigma}\), and Kirchhoff stress \(\mathbf{\tau}\). Further, each of the stress measures admits the same additive decomposition into equilibrium, non-equilibrium, and pressure components. For example, as will become clear in the dissipative structure, the Kirchhoff stress admits the expression,

\begin{equation}
\mathbf{\tau} = \mathbf{\tau}_{eq}^{(A)}\mathbf{+}\sum_{\ r \in \mathcal{R}_{neq}}^{}\mathbf{\tau}_{neq}^{(r\mathbf{)}} - p^{*}\mathbf{I}\mathbf{.}
\tag{2.6}\label{eq:2.6}
\end{equation}

\subsection*{2.3. Thermodynamic restriction and dissipative structure}

Under isothermal conditions, thermodynamic admissibility requires non-negative internal dissipation (\textbf{Eq. 1.4}). A standard way to enforce this structure is to write the inelastic flow in direction-magnitude form, an idea consistent with generalized-standard-material and associative finite-strain inelasticity \citep{Simo1992,Hackl1997}. In the finite-viscoelasticity literature, internal strain variables and their associated evolution equations have been formulated in the reference, intermediate, and current configurations by Le Tallec, Lion, and Reese--Govindjee, respectively \citep{LeTallec1993,Lion1997,Reese1998a,Gouhier2024}. The spatial velocity gradient is defined as

\begin{equation}
\mathbf{L =}\frac{\partial\dot{\mathbf{x}}}{\partial\mathbf{x}} = \dot{\mathbf{F}}\mathbf{F}^{- 1} = \mathbf{D} + \mathbf{W},
\tag{2.7}\label{eq:2.7}
\end{equation}

where \(\mathbf{D},\mathbf{W}\) are the symmetric and skew-symmetric parts. We can further decompose the rate of change of the viscous deformation in the relaxed (intermediate) configuration:

\begin{equation}
\mathbf{L}_{v}^{(r)} = {\dot{\mathbf{F}}}_{v}^{(r)}\mathbf{F}_{v}^{(r) - 1} = \mathbf{D}_{v}^{(r)} + \mathbf{W}_{v}^{(r)},
\tag{2.8}\label{eq:2.8}
\end{equation}

or also in the current configuration:

\begin{equation}
{\widetilde{\mathbf{L}}}_{v}^{(r)} = \mathbf{F}_{e}^{(r)}\mathbf{L}_{v}^{(r)}\mathbf{F}_{e}^{(r) - 1} = \mathbf{F}_{e}^{(r)}{\dot{\mathbf{F}}}_{v}^{(r)}\mathbf{F}_{v}^{(r) - 1}\mathbf{F}_{e}^{(r) - 1} = {\widetilde{\mathbf{D}}}_{v}^{(r)} + {\widetilde{\mathbf{W}}}_{v}^{(r)},
\tag{2.9}\label{eq:2.9}
\end{equation}

where \({\widetilde{\mathbf{D}}}_{v}^{(r)}\) and \({\widetilde{\mathbf{W}}}_{v}^{(r)}\) are the spatial viscous rate-of-deformation and spin tensors, respectively. The multiplicative decomposition of the deformation gradient gives

\begin{equation}
\mathbf{L} = \mathbf{L}_{e}^{(r)} + {\widetilde{\mathbf{L}}}_{v}^{(r)}\ \forall r,
\tag{2.10}\label{eq:2.10}
\end{equation}

where \(\mathbf{L}_{e}^{(r)} = {\dot{\mathbf{F}}}_{e}^{(r)}\mathbf{F}_{e}^{(r) - 1}.\) The multiplicative decomposition introduces, for each branch \(r\), a rotationally indeterminate relaxed intermediate configuration. Indeed, for any proper orthogonal tensor \(\mathbf{Q}^{(r)}:\)

\begin{equation}
\mathbf{F}_{e}^{(r)*} = \mathbf{F}_{e}^{(r)}\mathbf{Q}^{(r)T},\mathbf{F}_{v}^{(r)*} = \mathbf{Q}^{(r)}\mathbf{F}_{v}^{(r)},
\tag{2.11}\label{eq:2.11}
\end{equation}

while

\begin{equation}
\mathbf{F}_{e}^{(r)*}\mathbf{F}_{v}^{(r)*} = \mathbf{F}
\tag{2.12}\label{eq:2.12}
\end{equation}

The intermediate state can be made unique in different ways, as discussed by Boyce, Weber and Parks \citep{Boyce1989}. One convenient constitutive choice is a spatially irrotational viscous flow \({\widetilde{\mathbf{W}}}_{v}^{(r)} = 0\) and therefore, \({\widetilde{\mathbf{L}}}_{v}^{(r)} = {\widetilde{\mathbf{D}}}_{v}^{(r)}\). Using \textbf{Eq. 2.9}

\begin{equation}
{\dot{\mathbf{F}}}_{v}^{(r)} = \mathbf{F}_{e}^{(r) - 1}{\widetilde{\mathbf{D}}}_{v}^{(r)}\mathbf{F}_{e}^{(r)}\mathbf{F}_{v}^{(r)} = \mathbf{F}_{e}^{(r) - 1}{\widetilde{\mathbf{D}}}_{v}^{(r)}\mathbf{F.}
\tag{2.13}\label{eq:2.13}
\end{equation}

A key constitutive assumption is to write the branch flow rule in magnitude-direction form as

\begin{equation}
{\widetilde{\mathbf{D}}}_{v}^{(r)} = {\dot{\gamma}}^{(r)}\mathbf{N}^{(r)},
\tag{2.14}\label{eq:2.14}
\end{equation}

where \(\mathbf{N}^{(r)}\) is the normalized deviatoric stress direction

\begin{equation}
\mathbf{N}^{(r)} = \frac{\mathbf{\tau}_{dev}^{(r)}}{\parallel \mathbf{\tau}_{dev}^{(r)} \parallel},
\tag{2.15}\label{eq:2.15}
\end{equation}

where \(\parallel \mathbf{A} \parallel : = \left( \mathbf{A}:\mathbf{A} \right)^{\frac{1}{2}}\) denotes the norm, \(\mathbf{\tau}_{dev}^{(r)} = \mathbf{\tau}^{\mathbf{(}r)} - \frac{1}{3}tr\left( \mathbf{\tau}^{(r)} \right)\mathbf{I}\) is the deviatoric Kirchhoff stress, and \({\dot{\gamma}}^{(r)}\) is a scalar rate of viscous deformation. To complement this description, we introduce the objective scalar, the effective driving stress

\begin{equation}
\tau_{eff}^{(r)} = {\frac{1}{\sqrt{2}}\left( \text{\,}\mathbf{\tau}_{dev}^{(r)}:\mathbf{\tau}_{dev}^{(r)} \right)}^{1/2} = \frac{1}{\sqrt{2}} \parallel \mathbf{\tau}_{dev}^{(r)} \parallel ,
\tag{2.16}\label{eq:2.16}
\end{equation}

which represents an effective equivalent measure of the deviatoric driving stress. Equivalently, \(\tau_{eff}^{(r)} = \sqrt{J_{2}^{(r)}}\), where \(J_{2}^{(r)}\) is the second deviatoric stress invariant of \(\mathbf{\tau}_{dev}^{(r)}\). Hence, \(\mathbf{N}^{(r)}\) specifies the flow direction, while \(\tau_{eff}^{(r)}\) provides the scalar stress measure entering the branch evolution scalar law,

\begin{equation}
{\dot{\gamma}}^{(r)} = \widehat{{\dot{\gamma}}_{r}}\left( \tau_{eff}^{(r)},q^{(r)},Z \right).
\tag{2.17}\label{eq:2.17}
\end{equation}

Thus, constitutive modeling is reduced to the determination of the function \(\widehat{{\dot{\gamma}}_{r}}\) , which in general can depend on the effective driving stress \(\tau_{eff}^{(r)}\) as well as branch-specific objective internal-state measures \(q^{(r)}\) (such as \(I_{1,v}\)) as well as, when relevant, material descriptors \(Z\). With this construction, the branch-wise nonequilibrium dissipation becomes

\begin{equation}
\mathcal{D}_{int} = \sum_{r \in \mathcal{R}_{neq}}^{}\mathbf{\tau}_{dev}^{(r)}:{\widetilde{\mathbf{D}}}_{v}^{(r)} \geq 0.
\tag{2.18}\label{eq:2.18}
\end{equation}

The derivation of \textbf{Eq. 2.18} from the Clausius--Duhem inequality is further detailed in \textbf{Appendix B}. This dissipation inequality, together with the direction-scalar form in \textbf{Eq. 2.14}, results in a very simple constraint to ensure thermodynamic consistency:

\begin{equation}
{\dot{\gamma}}^{(r)} \geq 0.
\tag{2.19}\label{eq:2.19}
\end{equation}

\section*{3. Multi-material constitutive modeling}

\subsection*{3.1. Equilibrium branch}

The equilibrium branch defines the material\textquotesingle s quasi-static response. Rather than calibrating a separate equilibrium model for each composition, we condition the scalar constitutive functions on the descriptor vector \(Z\). In other words, \(Z\) controls how the response varies across related materials while preserving a common finite-strain constitutive model structure. Two alternatives are used for this branch: \emph{(i)} a \(Z\)-dependent Arruda--Boyce model, which provides an interpretable representation through composition-dependent chain network parameters, and \emph{(ii)} a \(Z\)-conditioned NODE model, which provides a more flexible learned representation of the scalar constitutive functions.

\subsubsection*{3.1.1 Closed-form multi-material conditioning of the hyperelastic response}

For the interpretable model, we choose the incompressible Arruda--Boyce (AB) eight-chain model \citep{Arruda1993}. For isotropic incompressible elasticity, the equilibrium response is expressed in terms of the first invariant of the left Cauchy--Green tensor

\begin{equation}
I_{1} = tr\left( \mathbf{B} \right),\mathbf{B} = \mathbf{F}\mathbf{F}^{T}
\tag{3.1}\label{eq:3.1}
\end{equation}

and the corresponding chain stretch is

\begin{equation}
\lambda_{\text{chain}} \coloneqq \sqrt{\frac{I_{1}}{3}}\ .
\tag{3.2}\label{eq:3.2}
\end{equation}

The equilibrium strain-energy density is taken as

\begin{equation}
\psi_{AB} = \mu\sqrt{N}\left\lbrack \lambda_{\text{chain}}\text{\,}\beta_{chain} + \sqrt{N}\ln\left( \frac{\beta_{chain}}{\sinh\beta_{chain}} \right) \right\rbrack + C,
\tag{3.3}\label{eq:3.3}
\end{equation}

where

\begin{equation}
\beta_{chain} \coloneqq \mathcal{L}^{- 1}\left( \chi_{chain} \right) \approx \frac{\chi_{chain}\left( 3 - {\chi_{chain}}^{2} \right)}{1 - {\chi_{chain}}^{2}},\ \ \chi_{chain} \coloneqq \frac{\lambda_{\text{chain}}}{\sqrt{N}}
\tag{3.4}\label{eq:3.4}
\end{equation}

\(\mathcal{L}^{- 1}\) denotes the inverse Langevin function, approximated here by the Padé rational form proposed by Cohen \citep{Cohen1991}. The admissible range is

\begin{equation}
0 \leq \chi_{chain} < 1,
\tag{3.5}\label{eq:3.5}
\end{equation}

since the Langevin function takes values in \(( - 1,1)\), while its inverse diverges as \(\chi_{chain} \rightarrow 1^{-}\). This asymptotic behavior encodes the finite extensibility of polymer chains and is responsible for the strain-stiffening response. To explicitly encode multi-material dependence, we introduce functions

\begin{equation}
\begin{gathered}
\mu(Z) = \mathcal{N}_{\mu}(Z),\ \\[3pt]
N(Z) = \mathcal{N}_{N}(Z).
\end{gathered}
\tag{3.6}\label{eq:3.6}
\end{equation}

Where \(\mathcal{N}_{\mu}(Z)\) and \(\mathcal{N}_{N}(Z)\) are neural networks (specifically multi-layer perceptrons or MLP) which take in the material feature vector \(Z\) and output the parameters of the AB model.

For benchmarking, we also consider other classical \(I_{1}\)-based closed-form hyperelastic models extended to the multi-material setting by allowing their constitutive parameters to depend on the conditioning vector \(Z\). In addition to Arruda--Boyce, we also do multi-material conditioning for neo-Hookean and Fung-type models for comparison (see \Tabref{1}). The neo-Hookean model provides the simplest baseline; its energy is linear in \(I_{1}\), its derivative (\(\psi_{1}: = \frac{\partial\psi}{\partial I_{1}}\)) is constant, and therefore it does not produce strain stiffening. The Fung--Demiray model \citep{Fung1967,Demiray1972} introduces exponential stiffening rather than the finite-extensibility threshold. The Arruda--Boyce is the micromechanics-based choice, in which strain stiffening arises from the finite extensibility of representative polymer chains, with \(\mu\) setting the small-strain stiffness and \(N\) controlling the chain extensibility. Following the strategy used for the Arruda--Boyce model in \textbf{Eq.} \textbf{3.6}, the parameters of the other closed-form models are also represented by neural networks conditioned on the descriptor vector \(Z\).

\hypertarget{tab:1}{\textbf{Table 1.}} Material-conditioned \(I_{1}\)-based equilibrium models

\begin{longtable}[]{@{}
  >{\raggedright\arraybackslash}p{(\columnwidth - 6\tabcolsep) * \real{0.2400}}
  >{\raggedright\arraybackslash}p{(\columnwidth - 6\tabcolsep) * \real{0.1347}}
  >{\raggedright\arraybackslash}p{(\columnwidth - 6\tabcolsep) * \real{0.1966}}
  >{\raggedright\arraybackslash}p{(\columnwidth - 6\tabcolsep) * \real{0.4288}}@{}}
\toprule\noalign{}
\begin{minipage}[b]{\linewidth}\raggedright
Model
\end{minipage} & \begin{minipage}[b]{\linewidth}\raggedright
\textbf{Neo-Hookean}

(Constant)
\end{minipage} & \begin{minipage}[b]{\linewidth}\raggedright
\textbf{Fung-Demiray}

(Exponential)
\end{minipage} & \begin{minipage}[b]{\linewidth}\raggedright
\textbf{Arruda-Boyce}

(Inverse Langevin)
\end{minipage} \\
\midrule\noalign{}
\endhead
\bottomrule\noalign{}
\endlastfoot
Parameters & \(c(Z)\) & \(c(Z),a(Z)\) & \(N(Z),\mu(Z)\) \\
\(\psi\) & \(c\left( I_{1} - 3 \right)\) & \(\frac{c}{2a}\left\lbrack e^{a(I_{1} - 3)} - 1 \right\rbrack\) & \(\mu\sqrt{N}\left\lbrack \lambda_{\text{chain}}\text{\,}\beta_{chain} + \sqrt{N}\ln\left( \frac{\beta_{chain}}{\sinh\beta_{chain}} \right) \right\rbrack + C\) \\
\(\psi_{1} \coloneqq \frac{\partial\psi}{\partial I_{1}}\) & \(c\) & \(\frac{c}{2}e^{a(I_{1} - 3)}
\) & \(\frac{\mu}{6}A\left( \chi_{chain} \right),\)

\(A\left( \chi_{chain} \right) \coloneqq \frac{\beta_{chain}}{\chi_{chain}} \approx \frac{\left( 3 - {\chi_{chain}}^{2} \right)}{\left( 1 - {\chi_{chain}}^{2} \right)}\) \\
\(\alpha \coloneqq 2\psi_{1}\) & \(2c\) & \({ce}^{a(I_{1} - 3)}\) & \(\frac{\mu}{3}A\left( \chi_{chain} \right)\) \\
\(G \coloneqq \alpha \mid_{I_{1} = 3}\)

Initial shear modulus & \(2c\) & \emph{c} & \(\frac{\mu}{3}A\left( \frac{1}{\sqrt{N}} \right) \approx \mu\text{\,}\frac{3N - 1}{3(N - 1)}\)

\(G \rightarrow \mu\) as \(N \rightarrow \infty\) \\
\end{longtable}

Finally, for incompressible materials with \(\psi = \psi\left( I_{1} \right)\), the corresponding Cauchy and first Piola--Kirchhoff stresses are,

\begin{equation}
\begin{gathered}
\mathbf{\sigma} = 2\frac{\partial\psi}{\partial I_{1}}\mathbf{B} - p^{*}\mathbf{1,} \\[3pt]
\bm{P} = \mathbf{\sigma}\mathbf{F}^{\mathbf{- T}}\mathbf{=}2\frac{\partial\psi}{\partial I_{1}}\mathbf{F} - p^{*}\mathbf{F}^{- T}.
\end{gathered}
\tag{3.7}\label{eq:3.7}
\end{equation}

For the uniaxial loading case in the 11-direction:

\begin{equation}
\begin{gathered}
\sigma_{11} = 2\text{\,}\frac{\partial\psi}{\partial I_{1}}\left( \lambda^{2}-\lambda^{- 1} \right) = {\lambda P}_{11}, \\[3pt]
\ P_{11} = 2\text{\,}\frac{\partial\psi}{\partial I_{1}}\left( \lambda-\lambda^{- 2} \right).
\end{gathered}
\tag{3.8}\label{eq:3.8}
\end{equation}

Therefore, when the equilibrium energy depends only on \(I_{1}\), the quantity that directly controls the equilibrium stress response is the derivative \(\frac{\partial\psi}{\partial I_{1}}\). Depending on the constitutive model, this term may be constant, exponential, or of a more general closed-form type. In the NODE case, similarly, the interpolation of the derivative \(\frac{\partial\psi}{\partial I_{1}}\) will be the core of the constitutive model. For convenience, we introduce the notation

\begin{equation}
\alpha \coloneqq 2\frac{\partial\psi}{\partial I_{1}},
\tag{3.9}\label{eq:3.9}
\end{equation}

which will be used throughout the remainder of the formulation.

\subsubsection*{3.1.2 Data-driven hyperelastic response based on NODEs}

In addition to closed-form hyperelastic models, the equilibrium response can also be represented through a neural ordinary differential equation (NODE) defined in terms of the invariant \(I_{1}\). For a fixed conditioning vector \(Z\), we introduce an initial state condition \(y(0) = I_{1} - 3\), as shown in \Figref[A]{2}, and evolve it over an artificial coordinate \(s \in \lbrack 0,1\rbrack\) to get a final value at \(y(1)\) according to

\begin{equation}
\frac{dy}{ds} = f_{\theta}(y,Z)\text{\,}y,s \in \lbrack 0,1\rbrack,
\tag{3.10}\label{eq:3.10}
\end{equation}

here, \(f_{\theta}(y,Z)\) is a scalar learned function represented by a neural network (MLP) with parameters \(\theta\).

The NODE is evaluated numerically by discretizing the artificial coordinate \(s\) using an explicit Euler solver \(y_{j + 1} = y_{j} + \Delta s\text{\,}f_{\theta}\left( y_{j},Z \right)y_{j}\), where denotes the \(j\)-th discretization step along the artificial coordinate \(s\). The multiplicative form makes \(y = 0\) an invariant solution and preserves the sign of the continuous flow, \(y(0) \geq 0\  \Rightarrow \ y(s) \geq 0.\) Moreover, uniqueness of the ODE flow prevents trajectories initialized at different values from crossing, since any intersection would imply identical subsequent solutions, thereby preserving their ordering and making the terminal map \(y(1)\) non-decreasing with respect to \(I_{1}\). The final state \(y(1)\) is then used to define the elastic energy gradient as

\begin{equation}
\psi_{1} = \frac{\partial\psi}{\partial I_{1}} = y(1) + softplus\left( E_{\theta}(Z) \right),
\tag{3.11}\label{eq:3.11}
\end{equation}

where \(E_{\theta}(Z)\) is a second learned positive scalar function (another MLP) depending only on the conditioning vector \(Z\), and \(softplus(x) = ln(1 + e^{x})\) is used to ensure positivity. This construction is attractive for several reasons. First, the undeformed state corresponds to \(I_{1} = 3\), for which \(y(0) = 0\), and the multiplicative form of the ODE gives \(y(s) \equiv 0\). As a result,

\begin{equation}
{\frac{\partial\psi}{\partial I_{1}} \mid}_{I_{1} = 3} = {softplus}\left( E_{\theta}(Z) \right).
\tag{3.12}\label{eq:3.12}
\end{equation}

This directly determines the \(Z\)-dependent initial value of \(\frac{\partial\psi}{\partial I_{1}}\), and therefore the corresponding small-strain shear modulus \(G\),

\begin{equation}
G_{NODE}(Z) = 2{\frac{\partial\psi}{\partial I_{1}} \mid}_{I_{1} = 3} = 2\text{\,}{softplus}{\left( E_{\theta}(Z) \right).}
\tag{3.13}\label{eq:3.13}
\end{equation}

Second, for admissible uniaxial states with \(I_{1} \geq 3\), the solution remains non-negative, while the additional softplus offset ensures \(\frac{\partial\psi}{\partial I_{1}} \geq 0\). The monotonicity and non-negativity of \(\frac{\partial\psi}{\partial I_{1}}\) ensures that \(\psi\) is convex and monotonic on \(I_{1}\) and thus polyconvex. Considering the response in terms of small-strain shear and correction at larger strains, we introduce also the notation (see \Suppref{2}{Table~S1} in the Supplementary Material),

\begin{equation}
\frac{\partial\psi}{\partial I_{1}} = \frac{G}{2} + \Delta\psi_{1}\left( I_{1} \right),
\tag{3.14}\label{eq:3.14}
\end{equation}

where \(\Delta\psi_{1} = y\left( s = 1;\ y(0) = I_{1} - 3 \right)\) represents the deformation-dependent strain-stiffening contribution. In the NODE formulation, these two contributions are learned separately: the initial shear modulus \(G\) is learned as an offset to the NODE trajectories, whereas the NODE evolution determines the strain-stiffening response. In contrast, in closed-form models these two contributions are coupled through their prescribed functional form and constitutive parameters.

\subsection*{3.2. Nonequilibrium branch}

The nonequilibrium response is commonly interpreted as a superposition of relaxation mechanisms acting over distinct deformation and time scales. However, in digital materials, it is difficult to assert \emph{a priori} that each branch represents a single, uniquely identifiable molecular mechanism due to the combined effects of multiple unobserved microscopic processes whose characteristic timescales may overlap and whose individual contributions cannot be isolated from the available mechanical data alone. We therefore consider as few branches as possible to describe the data accurately and robustly, and interpret the branches as effective constitutive modes. Each branch is characterized by two scalar constitutive ingredients: a stored-energy function defined by its own elastic state, and a non-negative evolution law governing dissipation.

\subsubsection*{3.2.1. Data-driven and closed-form modeling of the stored elastic energy in non-equilibrium branches}

Each nonequilibrium branch stores energy through its own elastic state. Here, \(r\  \in \ \left\{ B,C \right\}\) denotes the \(r\)-th nonequilibrium branch. For the isotropic incompressible case adopted here, the branch free energy is restricted to depend only on the elastic first isochoric invariant,

\begin{equation}
\begin{gathered}
\psi^{(r)} = {\widehat{\psi}}^{(r)}\left( I_{1,e}^{(r)} \right), \\[3pt]
I_{1,e}^{(r)} = tr\left( \mathbf{B}_{e}^{(r)} \right).
\end{gathered}
\tag{3.15}\label{eq:3.15}
\end{equation}

This induces the scalar model, analogous to that defined in \textbf{Eq. 3.9},

\begin{equation}
\alpha^{(r)}\left( I_{1,e}^{(r)};Z \right) \coloneqq 2\text{\,}\frac{\partial\psi^{(r)}}{\partial I_{1,e}^{(r)}},
\tag{3.16}\label{eq:3.16}
\end{equation}

where the explicit dependence on the material features \(Z\) has been introduced. It is convenient to restate the Kirchhoff stress from \textbf{Eq. 2.6} in terms of the elastic deformation of the branch rather than the push-forward of the nominal stress, namely

\begin{equation}
\mathbf{\tau}_{neq}^{(r)} = 2\frac{\partial\psi^{(r)}}{\partial\mathbf{B}_{e}^{(r)}}\mathbf{B}_{e}^{(r)} = \alpha^{(r)}\mathbf{B}_{e}^{(r)}.
\tag{3.17}\label{eq:3.17}
\end{equation}

Equation (\textbf{3.17}) shows that \(\alpha^{(r)}\) summarizes the constitutive model into one scalar function. In the nonequilibrium branches, \(\alpha^{(r)}\) is specified either in closed form or through a neural network representation, similar to what is described for the equilibrium branch. In the closed-form case, we restrict our attention to the Arruda--Boyce-type (AB) elastic representation. In the data-driven case, \(\alpha^{(r)}\) is represented through a NODE. Both representations are analogous to those introduced for the equilibrium response, the only difference being that they depend on the branch elastic invariant \(I_{1,e}^{(r)}\) rather than on the total invariant \(I_{1}\):

\begin{equation}
\alpha^{(r)}(I_{1,e}^{(r)})\dot{=}\left\{ \begin{array}{r}
AB:\ \frac{\mu^{(r)}(Z)}{3}\frac{\sqrt{N^{(r)}(Z)}}{\lambda_{chain}^{e,(r)}}\mathcal{L}^{- 1}\left( \frac{\lambda_{chain}^{e,(r)}}{\sqrt{N^{(r)}(Z)}} \right) \\
NODE:\ \left\{ \begin{matrix}
2\left\lbrack y^{(r)}(1) + softplus\left( E_{\theta}^{(r)}(Z) \right) \right\rbrack \\
{\dot{y}}^{(r)} = {f_{\theta}}^{(r)}(y^{(r)},Z)\text{\,}y^{(r)} \\
y^{(r)}(0) = I_{1,e}^{(r)} - 3
\end{matrix} \right.\ 
\end{array} \right.\ .
\tag{3.18}\label{eq:3.18}
\end{equation}

To clarify the constitutive structure of each nonequilibrium branch in this work, \Figref[B]{2} highlights two interacting lanes of the constitutive model. The left lane, which we term tensorial operations, goes from the internal memory tensor \(\mathbf{F}_{v}^{(r)}\) and ends at the viscous rate-of-deformation tensor \({\widetilde{\mathbf{D}}}_{v}^{(r)}\), through intermediate kinematic and stress tensors. The right lane is the constitutive response, which is boiled down to two scalar functions: mapping the material descriptor \(Z\) and branch invariant \(I_{1,e}^{(r)}\) into the energy-related coefficient \(\alpha^{(r)}\), and the effective driving stress \(\tau_{eff}^{(r)}\) and material descriptor \(Z\) into the dissipation rate \({\dot{\gamma}}^{(r)}\). The choice of kinematic split and dissipation along the non-equilibrium stress direction should not be interpreted as unique, but they are physically interpretable representations of the constitutive framework such that scalar artificial neural networks with minimal requirements (monotonicity and non-negativity) provide an extremely rich space to describe the materials described by features \(Z\) while automatically satisfying thermodynamic and mathematical requirements.

\begin{figure}[htbp]
\centering
\includegraphics[width=\linewidth,keepaspectratio]{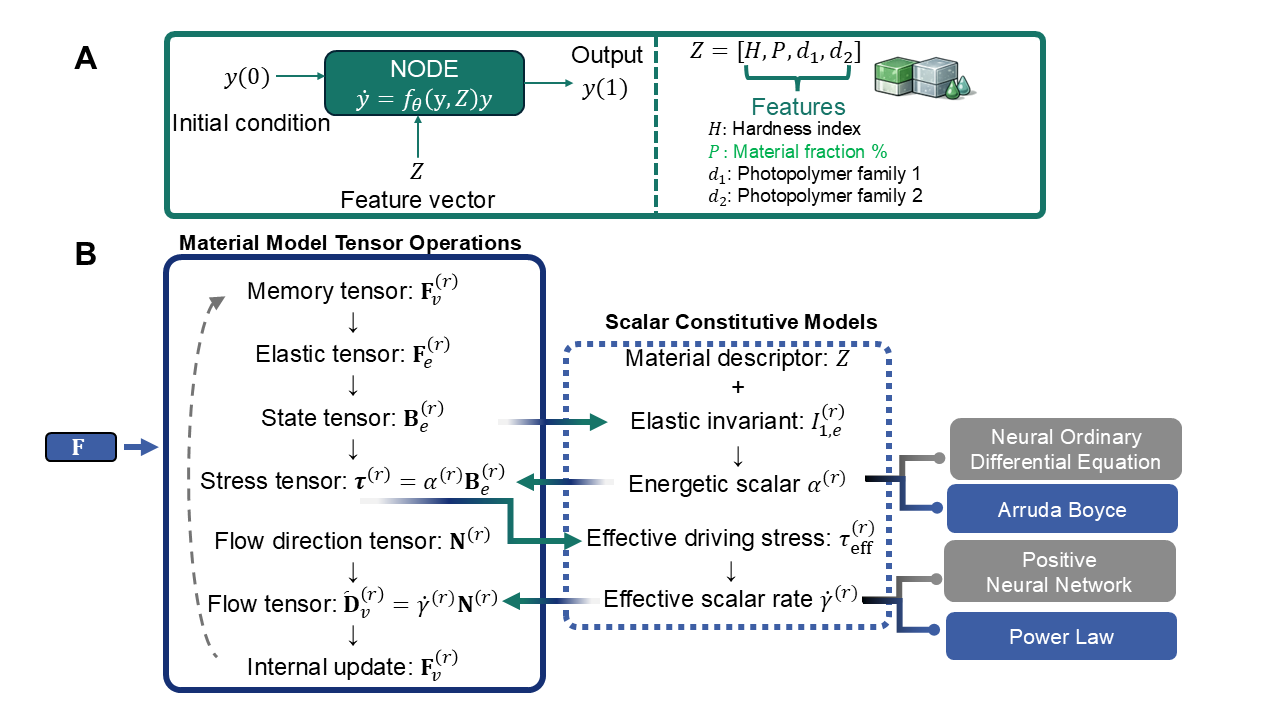}
\caption{\textbf{(A)} NODE-based conditioning of the constitutive response by the material descriptor vector \(Z = \lbrack H,P,d_{1},d_{2}\rbrack\). \textbf{(B)} Schematic of the dissipative flow structure of a nonequilibrium branch, from current internal memory to viscous update going through the driving stress. All the tensorial calculations are isolated from the constitutive model functions which are scalar: the strain energy of the branch is captured in \(\alpha^{(r)}\) and the dissipation rate in\(\ {\dot{\gamma}}^{(r)}\).}
\label{fig:2}
\end{figure}

\subsubsection*{3.2.2. Closed-form and data-driven branch dissipation}

A classical Bergström--Boyce-type branch kinetics used the principles of the Doi--Edwards reptation mode for the dissipation rate \citep{Doi1988}, and may be written as a non-negative overstress law of the form

\begin{equation}
{\dot{\gamma}}^{(r)}(\tau_{eff}^{(r)},\ \lambda_{v,chain}^{(r)}) = {\dot{\gamma}}_{0}{\times \left( \lambda_{v,chain}^{(r)}-1 \right)}^{C} \times \left( \frac{\tau_{eff}^{(r)}}{\tau_{ref}} \right)^{m},
\tag{3.19}\label{eq:3.19}
\end{equation}

where \(C\) and \(m\) are material parameters \citep{Bergstrm1998}. Here, \(\tau_{ref}\) is a reference stress introduced for dimensional consistency and therefore has the same units as \(\tau_{eff}^{(r)}\), while \({\dot{\gamma}}_{0}\) is a reference rate. In this form, the branch flow rate is non-negative by construction. In the formulation discussed by Dal and Kaliske, thermodynamic consistency is satisfied for \({\dot{\gamma}}_{0}/{(\tau_{ref})}^{\text{\,}m} > 0\) and \(m > 0\) \citep{Dal2009}.

We generalize the form in \textbf{Eq. 3.19} and also incorporate the multi-material extension, i.e. parameter learning as a function of material features \(Z\) through neural networks (MLPs). We considered the following generalized power law form:

\begin{equation}
\begin{gathered}
{\dot{\gamma}}^{(r)}(\tau_{eff}^{(r)},\lambda_{v,chain}^{(r)},Z) = K_{1}^{(r)}(Z)\times\text{\,}\frac{\tau_{eff}^{(r)}}{f_{1}^{(r)}\left( u^{(r)} \right)} + K_{2}^{(r)}(Z) \times \left( \frac{\tau_{eff}^{(r)}}{f_{2}^{(r)}\left( u^{(r)} \right)} \right)^{m_{3}^{(r)}(Z)}, \\[3pt]
u^{(r)} = \mid \lambda_{v,chain}^{(r)} - 1 \mid .
\end{gathered}
\tag{3.20}\label{eq:3.20}
\end{equation}

Although, in principle, \(K_{1}^{(r)}\) and \(f_{1}^{(r)}\) could be combined into a single equivalent function \(f(u^{(r)},Z)\), it is preferable to keep the material-dependent and memory-dependent contributions separated within the neural-network parameterization to mitigate gradient cancellation and parameter confounding during training (similar case for \(K_{2}^{(r)}\) and \(f_{2}^{(r)}\)). This choice defines the baseline generalized form, from which alternative model variants are later introduced to explore different power law simplifications and assumptions. In addition, to generalize the branch kinetics while preserving the same thermodynamic structure, we can replace the closed-form scalar evolution law with a learned non-negative function,

\begin{equation}
{\dot{\gamma}}_{X} = \tau_{eff}^{(r)}\times\phi_{\theta}^{(r)}(\tau_{eff}^{(r)},q^{(r)},Z),
\tag{3.21}\label{eq:3.21}
\end{equation}

where,

\begin{equation}
\phi_{\theta}^{(r)} = softplus({\phi^{*}}_{\theta}^{(r)}(\tau_{eff}^{(r)},q^{(r)},Z)) \geq 0,
\tag{3.22}\label{eq:3.22}
\end{equation}

here \({\phi^{*}}_{\theta}^{(r)}\) is a standard MLP, while the softplus activation ensures \(\phi_{\theta}^{(r)} \geq 0\) by construction and \(\tau_{eff}^{(r)} \geq 0\), the branch flow remains consistent with the dissipative structure introduced in \textbf{Eq.} \textbf{2.18}. Here, the memory internal measure \(q^{(r)}\) is replaced by the viscous branch invariants \(I_{1,v}^{(r)}\) (equivalent representations, such as \(\lambda_{v,chain}^{(r)}\), can also be used). The multiplicative form of \textbf{Eq. 3.21} ensures that zero overstress produces zero flow, while \(\phi_{\theta}^{(r)} \geq 0\) guarantees \({\dot{\gamma}}^{(r)} \geq 0\) and non-negative dissipation. In summary, the two classes of branch-flow functions considered are (see also \Figref[B]{2}):

\begin{equation}
{\dot{\gamma}}^{(r)}\left( \tau_{eff}^{(r)},q^{(r)},Z \right)\dot{=}\left\{ \begin{array}{r}
Power\ Law:K_{1}^{(r)}(Z)\times\text{\,}\frac{\tau_{eff}^{(r)}}{f_{1}^{(r)}\left( u^{(r)} \right)} + K_{2}^{(r)}(Z) \times \left( \frac{\tau_{eff}^{(r)}}{f_{2}^{(r)}\left( u^{(r)} \right)} \right)^{m_{3}^{(r)}(Z)} \\
Neural\ Network:\tau_{eff}^{(r)}\phi_{\theta}^{(r)},\phi_{\theta}^{(r)} \geq 0
\end{array}. \right.\
\tag{3.23}\label{eq:3.23}
\end{equation}

\section*{4. Uniaxial incompressible case}

For calibration against homogeneous uniaxial tests, the three-dimensional formulation specializes to incompressible uniaxial deformation. The total deformation gradient is written as

\begin{equation}
\mathbf{F} = diag(\lambda,\lambda^{- 1/2},\lambda^{- 1/2}),
\tag{4.1}\label{eq:4.1}
\end{equation}

so that \(\det\mathbf{F} = 1\). Because the nonequilibrium branch evolution is driven by deviatoric stress, the viscous distortion is taken to be isochoric, \(\det\mathbf{F}_{v}^{(r)} = 1\). The viscous distortion of each nonequilibrium branch is therefore parametrized as

\begin{equation}
\mathbf{F}_{v}^{(r)} = diag\left( \lambda_{v}^{(r)},(\lambda_{v}^{(r)})^{- 1/2},(\lambda_{v}^{(r)})^{- 1/2} \right).
\tag{4.2}\label{eq:4.2}
\end{equation}

Together with the multiplicative decomposition \(\mathbf{F} = \mathbf{F}_{e}^{(r)}\mathbf{F}_{v}^{(r)}\)and \(\det\mathbf{F} = 1\), this gives \(\det\mathbf{F}_{e}^{(r)} = 1\). The branch elastic stretch is then

\begin{equation}
\lambda_{e}^{(r)} = \frac{\lambda}{\lambda_{v}^{(r)}},
\tag{4.3}\label{eq:4.3}
\end{equation}

and the corresponding elastic first invariant becomes

\begin{equation}
I_{1,e}^{(r)} = (\lambda_{e}^{(r)})^{2} + 2(\lambda_{e}^{(r)})^{- 1}.
\tag{4.4}\label{eq:4.4}
\end{equation}

In this case, the non-negative scalar effective branch driving stress reduces to

\begin{equation}
\tau_{eff}^{(r)} = \frac{\alpha^{(r)}}{\sqrt{3}} \mid (\lambda_{e}^{(r)})^{2} - (\lambda_{e}^{(r)})^{- 1} \mid .
\tag{4.5}\label{eq:4.5}
\end{equation}

For~later~use~in~the~analysis, it is convenient to also define the corresponding signed quantity,

\begin{equation}
{\widetilde{\tau}}_{eff}^{(r)} = \frac{\alpha^{(r)}}{\sqrt{3}}\lbrack(\lambda_{e}^{(r)})^{2} - (\lambda_{e}^{(r)})^{- 1}\rbrack.
\tag{4.6}\label{eq:4.6}
\end{equation}

The evolution of the internal viscous stretch can be written in scalar form as

\begin{equation}
{\dot{\lambda}}_{v}^{(r)} = \sqrt{\frac{2}{3}}\text{\,}{\dot{\gamma}}^{(r)}\text{\,}\lambda_{v}^{(r)}\text{\,}sign\left( (\lambda_{e}^{(r)})^{2} - (\lambda_{e}^{(r)})^{- 1} \right) = \lambda_{v}^{(r)}{\widetilde{D}}_{v}^{11,\ (r)},
\tag{4.7}\label{eq:4.7}
\end{equation}

where \({\widetilde{D}}_{v}^{11,\ (r)}\) is the 11-component of \({\widetilde{\mathbf{D}}}_{v}^{(r)}\) and \({\dot{\lambda}}_{v}^{(r)}\) the 11-component of \({\dot{\mathbf{F}}}_{v}^{(r)}\).The explicit uniaxial forms of \(\mathbf{B}_{e}^{(r)}\), \(I_{1,e}^{(r)}\), the corresponding deviatoric branch stress, effective driving stress, and normalized flow direction are derived in \textbf{Appendix C.3}. Combining the branch-wise dissipation structure in \textbf{Eq. 2.18} with its uniaxial specialization yields the resulting evolution equation, as detailed in \textbf{Appendix B.3}.

\begin{equation}
\mathcal{D}_{int} = \sum_{r \in \mathcal{R}_{neq}}^{}{\sqrt{2}\text{\,}\tau_{eff}^{(r)}{\dot{\gamma}}^{(r)}} \geq 0.
\tag{4.8}\label{eq:4.8}
\end{equation}

\section*{5. Materials and experimental data}

\subsection*{5.1 Base materials}

All specimens were 3D-printed on a Stratasys PolyJet J826 multi-material printer (Eden Prairie, MN, USA), which jet liquid photopolymers voxel by voxel, curing via ultraviolet. Two base resins were used: Agilus 30 Black, a soft elastomer, and Vero Ultra White, a stiff glassy polymer. Mixing the two in prescribed ratios during printing yields the manufacturer's digital materials, whose nominal Shore A labels rise with Vero content. Five of these materials were used here: the neat elastomer A30 and the digital mixtures A50, A70, A85, and A95. Their room-temperature behavior changes steadily across this range. A30 is compliant and recovers most of the applied deformation, whereas A95 is considerably stiffer and dissipates more energy.

Parts were printed in High Mix mode at a nominal 27 \(\mu\)m layer thickness. Each geometry was printed in a single fixed orientation, for consistency across materials. Support material was then removed mechanically, and the specimens were rinsed with water and air-dried before testing. \Tabref{2} lists the estimated Vero volume fraction \(\varphi_{v}\) of each material \citep{Slesarenko2018,Zorzetto2020}. The composition descriptor used throughout the constitutive model is the corresponding Agilus Black percentage, defined as \(P = 100 - \varphi_{v}\). Fabrication and characterization of this material family are described in \citep{Shen2026}.

\hypertarget{tab:2}{\textbf{Table 2.}} Digital materials used in this study, ordered by increasing Vero content. All five are voxel-scale digital mixtures of Agilus 30 Black and Vero Ultra White.

\begin{longtable}[]{@{}
  >{\raggedright\arraybackslash}p{(\columnwidth - 6\tabcolsep) * \real{0.0833}}
  >{\raggedright\arraybackslash}p{(\columnwidth - 6\tabcolsep) * \real{0.2500}}
  >{\raggedright\arraybackslash}p{(\columnwidth - 6\tabcolsep) * \real{0.3000}}
  >{\raggedright\arraybackslash}p{(\columnwidth - 6\tabcolsep) * \real{0.3667}}@{}}
\toprule\noalign{}
\begin{minipage}[b]{\linewidth}\raggedright
\textbf{\#}
\end{minipage} & \begin{minipage}[b]{\linewidth}\raggedright
\textbf{Material}
\end{minipage} & \begin{minipage}[b]{\linewidth}\raggedright
\textbf{Nominal Shore A}
\end{minipage} & \begin{minipage}[b]{\linewidth}\raggedright
\textbf{Vero volume fraction} \(\mathbf{\varphi}_{\mathbf{v}}\) \textbf{(\%)}
\end{minipage} \\
\midrule\noalign{}
\endhead
\bottomrule\noalign{}
\endlastfoot
1 & A30 & 30 & 0 \\
2 & A50 & 50 & 8.3 \\
3 & A70 & 70 & 18.1 \\
4 & A85 & 85 & 25.3 \\
5 & A95 & 95 & 36.2 \\
\end{longtable}

\subsection*{5.2 Mechanical testing}

Quasi-static uniaxial compression tests were performed at room temperature under displacement control on a ZwickRoell testing machine fitted with a 10 kN load cell (ZwickRoell GmbH \& Co. KG, Ulm, Germany). Cylindrical specimens were printed upright, with the cylinder axis along the layer-stacking direction Z. Those of A30, A50, and A70 measured 10 mm in diameter and height; those of the stiffer A85 and A95 measured 6 mm, which kept reaction forces within the load-cell capacity. Each specimen was compressed between hardened steel platens lubricated with a thin oil layer, which reduced friction and kept the deformation close to homogeneous. Displacement was measured optically from reference marks on the platens, so that machine compliance did not enter the recorded response.

Each test consisted of a single load--unload cycle to a maximum engineering strain of about 67\%, corresponding to a true strain of about 1.1. Four nominal engineering strain rates were used: 0.001, 0.01, 0.1, and 0.5 s\(^{-1}\). A fresh specimen was used in each test, and most rates were repeated to assess reproducibility. Before conversion, the force and displacement records were lightly smoothed to remove quantization noise from the optical tracking. Engineering stress was obtained from the measured force and the initial cross-sectional area, and engineering strain from the tracked displacement and the initial specimen height. True strain was taken as \(ln\lambda\), where \(\lambda\) is the ratio of current to initial height. Assuming incompressibility, true stress equals the engineering stress multiplied by \(\lambda\). Initial processing was carried out in MATLAB (MathWorks, Natick, MA, USA) and the conversion with a custom Python script (Python 3.14). The resulting stress--true strain curves form the data used for model training. Further details of the experimental procedure is given in \citep{Shen2026}.

\section*{6. Model training}

\subsection*{6.1 Equilibrium branch training}

The equilibrium branch was calibrated first using the quasi-equilibrium response extracted from the lowest-rate loading--unloading data. The approximation is most reliable for the more compliant compositions, where the lowest-rate hysteresis is small, and less exact for A70--A95, where a non-negligible nonequilibrium contribution may remain even at \(0.001\text{\,}s^{- 1}\). Therefore, the extracted curve is interpreted as a consistent way of approximating the equilibrium response rather than as a direct measurement of the fully relaxed response.

The data were randomly shuffled and optimized in mini-batches using the Adam optimizer with a learning rate of \(\eta_{A} = 10^{- 5}\) for \(N_{A} = 4000\) epochs, minimizing the mean-squared error between the predicted and experimental nominal stresses. For each data point, the stretch \(\lambda\) determines \(I_{1} = \lambda^{2} + 2\lambda^{- 1}\), together with the material descriptor \(Z\)\textbf{,} \(\alpha^{(A)}\)and \(P_{11}^{(A)}\) are evaluated using either the material-conditioned model, the closed-form model, or the NODE representation. After convergence, the equilibrium-branch parameters were frozen, and \(P_{11}^{(A)}\) was retained as the fixed equilibrium contribution during the training of branches B and C. The architecture and training settings are summarized in \Tabref{3}, and detailed training pseudocode is provided in \textbf{Appendix D.1}.

\subsection*{6.2 Nonequilibrium branch training}

For a given experimental curve \(i\) we have the measured stretch \(\lambda(t_{k})\), time \(t_{k}\), measured nominal stress \(P_{11,data}^{(i)}(t_{k})\), and the static material descriptor \(Z_{i}\) (embedded once per curve and reused at all time steps). For a given \(Z_{i}\), the model prediction is obtained by stepping sequentially through the measured history and updating the internal viscous stretches of the nonequilibrium branches. At time step \(k\), the inputs are \((\lambda_{k}\), \(t_{k})\), the internal states are (\(\lambda_{v,B}^{k},\lambda_{v,C}^{k}\)). For each branch \(r \in \{ B,C\}\), the one-dimensional update used in the present implementation proceeds as follows from \textbf{Eq.} \textbf{4.7}.

\begin{equation}
\lambda_{v,r}^{k + 1} = \lambda_{v,r}^{k} + \Delta t_{k}\sqrt{\frac{2}{3}}\text{\,}{\dot{\gamma}}_{r}^{k}\text{\,}\lambda_{v,r}^{k}\text{\,}sign(s_{r}^{k}).
\tag{5.1}\label{eq:5.1}
\end{equation}

For each curve \(i\), the prediction error is measured through a normalized root-mean-square error.

\begin{equation}
\begin{gathered}
{rng}_{i} = \underset{k}{\max}P_{11,data}^{(i)}(t_{k}) - \underset{k}{\min}P_{11,data}^{(i)}(t_{k}), \\[3pt]
{RMSE}_{i} = \left\lbrack \frac{1}{n_{i}}\sum_{k = 1}^{n_{i}}\left( P_{11,model}^{(i)}(t_{k}) - P_{11,data}^{(i)}(t_{k}) \right)^{2} \right\rbrack^{1/2}, \\[3pt]
L_{i} = \frac{{RMSE}_{i}}{{rng}_{i}},
\end{gathered}
\tag{5.2}\label{eq:5.2}
\end{equation}

\[
\]where \(n_{i}\) is the number of sampled time steps for curve \(i\). The normalization factor \({rng}_{i}\) is the experimental stress range of each curve, which prevents high-stress curves from dominating the loss. The training loss for one epoch is taken as the average over curves,

\begin{equation}
{\bar{\mathcal{L}}}_{epoch} = \frac{1}{N}\sum_{i = 1}^{N}L_{i},
\tag{5.3}\label{eq:5.3}
\end{equation}

so that each experimental history contributes with equal weight, independently of its absolute stress scale or number of samples. The optimization strategy, which uses a random permutation of loading curves (for different strain rates) is further discussed in \textbf{Appendix D.2}. Additionally, other details of the internal-state integration scheme, regularization, and the corresponding nonequilibrium-training pseudocode are also presented in \textbf{Appendix D.2}. The network architectures and training settings for Branches B and C are summarized in \Tabref{3}.

\hypertarget{tab:3}{\textbf{Table 3.}} Architecture and training settings of the multi-material equilibrium and nonequilibrium branches. Network architectures are written as input \(\rightarrow\) hidden layer(s) \(\rightarrow\) output. For example, 7 \(\rightarrow\) 32 \(\rightarrow\) 32 \(\rightarrow\) 1 denotes 7 input features, two hidden layers with 32 neurons each, and one output neuron.

\begin{longtable}[]{@{}
  >{\raggedright\arraybackslash}p{(\columnwidth - 4\tabcolsep) * \real{0.2673}}
  >{\raggedright\arraybackslash}p{(\columnwidth - 4\tabcolsep) * \real{0.3386}}
  >{\raggedright\arraybackslash}p{(\columnwidth - 4\tabcolsep) * \real{0.3941}}@{}}
\toprule\noalign{}
\begin{minipage}[b]{\linewidth}\raggedright
\textbf{Item}
\end{minipage} & \begin{minipage}[b]{\linewidth}\raggedright
\textbf{Equilibrium Branch A}
\end{minipage} & \begin{minipage}[b]{\linewidth}\raggedright
\textbf{Nonequilibrium Branches B and C}
\end{minipage} \\
\midrule\noalign{}
\endhead
\bottomrule\noalign{}
\endlastfoot
NODE field network \(f_{\theta}\) & MLP, 2\(\rightarrow 32 \rightarrow 32 \rightarrow 1\)

Inputs: \(P\) and \(I_{1}\) & MLP, 2\(\rightarrow 32 \rightarrow 32 \rightarrow 1\)

Inputs: \(P\) and \(I_{1}\) \\
Offset network \(E_{\theta}\) & MLP, 1\(\rightarrow 32 \rightarrow 1\)

Input: \(P\) & MLP, 1\(\rightarrow 32 \rightarrow 32 \rightarrow 1\)

Input: \(P\) \\
Flow network \(\phi_{\theta}^{*}\) & Not used & MLP, 3\(\rightarrow 32 \rightarrow 32 \rightarrow 1\)

Inputs: \(\tau_{eff}^{(r)},\mspace{6mu} I_{1,v}^{(r)},\mspace{6mu}\)and \(P\) \\
Hidden activation & \multicolumn{2}{>{\raggedright\arraybackslash}p{(\columnwidth - 4\tabcolsep) * \real{0.7327} + 2\tabcolsep}@{}}{%
\(\tanh\)} \\
NODE solver & \multicolumn{2}{>{\raggedright\arraybackslash}p{(\columnwidth - 4\tabcolsep) * \real{0.7327} + 2\tabcolsep}@{}}{%
Explicit Euler} \\
Euler steps & 16 & 16 \\
Loss & \(MSE\) & \(RMSE/rng\) \\
Optimizer & \multicolumn{2}{>{\raggedright\arraybackslash}p{(\columnwidth - 4\tabcolsep) * \real{0.7327} + 2\tabcolsep}@{}}{%
Adam} \\
Learning rate & \(10^{- 5}\) & \(10^{- 5}\) \\
Number of epochs & \(N_{epochs,A} = 4000\) & \(N_{epochs,BC} = 2500\) \\
Substepping & Not used & \(\Delta t_{\max} = 0.4\text{\,}s\), \(N_{sub,\max} = 16\) \\
Precision & \multicolumn{2}{>{\raggedright\arraybackslash}p{(\columnwidth - 4\tabcolsep) * \real{0.7327} + 2\tabcolsep}@{}}{%
float64} \\
\end{longtable}

\section*{7. Results and discussion}

\subsection*{7.1. Equilibrium response across multiple materials}

We first isolate the approximate equilibrium response and study how the multi-material framework describes the response across composition. Two model classes are compared: closed-form constitutive laws with learned material-dependent parameters, and data-driven NODE model. This comparison serves two purposes. First, it determines whether a shared multi-material parameterization is sufficient to capture the equilibrium hyperelastic response across compositions. Second, it clarifies the trade-off between interpretability and flexibility.

We show that the first invariant \(I_{1}\) is sufficient to capture the main nonlinearities and the systematic variation across materials via the NODE approach. The key point is that the model does not assume a fixed closed-form energy \(\psi(I_{1})\); instead, it learns a family of continuous hidden-state flows from the input \(y(0) = I_{1} - 3\), conditioned by the material descriptor \(Z\) (\Figref[A]{3}). In \Figref[B]{3}, the hidden-state NODE trajectories demonstrate that the flow in \(s\) depends on the material descriptor \(Z\) and can show very different qualitative behavior. For example, the learned transformation may either reduce or amplify the initial state \(y_{0} = I_{1} - 3\) depending on \(Z\). Because this is a flow with trajectories that do not cross, \(\partial\psi/\partial I_{1}\) is non-negative and monotone in \(I_{1}\) (see \Figref[C]{3}), which implies a monotonic increasing energy \(\psi\) and non-negative \(\partial^{2}\psi/\partial I_{1}^{2}\).

Notice that the offset term \(E_{\theta}(Z)\) depends only on the material descriptor and not on deformation, so it represents a composition-dependent initial shear modulus. Since \(P_{11}\) is directly controlled by \(\partial\psi/\partial I_{1}\) , once this function is specified, the framework outputs material response across the entire dataset as shown in \Figref[D]{3}. Supplementary \Suppref{17}{Fig.~S1} shows numerical differentiation, and integration allows us to recover \(\psi\) and \(\partial^{2}\psi/\partial I_{1}^{2}\) to illustrate the convexity of the energy. Also, the residuals remain tightly centered around zero over most of the stretch range, while the residual-versus-\(\lambda\) and Q--Q plots indicate that the remaining errors are not purely Gaussian noise, but are instead concentrated in a small subset of more challenging deformation and material regions (see \Suppref{18}{Fig.~S2}).

\begin{figure}[htbp]
\centering
\includegraphics[width=\linewidth,keepaspectratio]{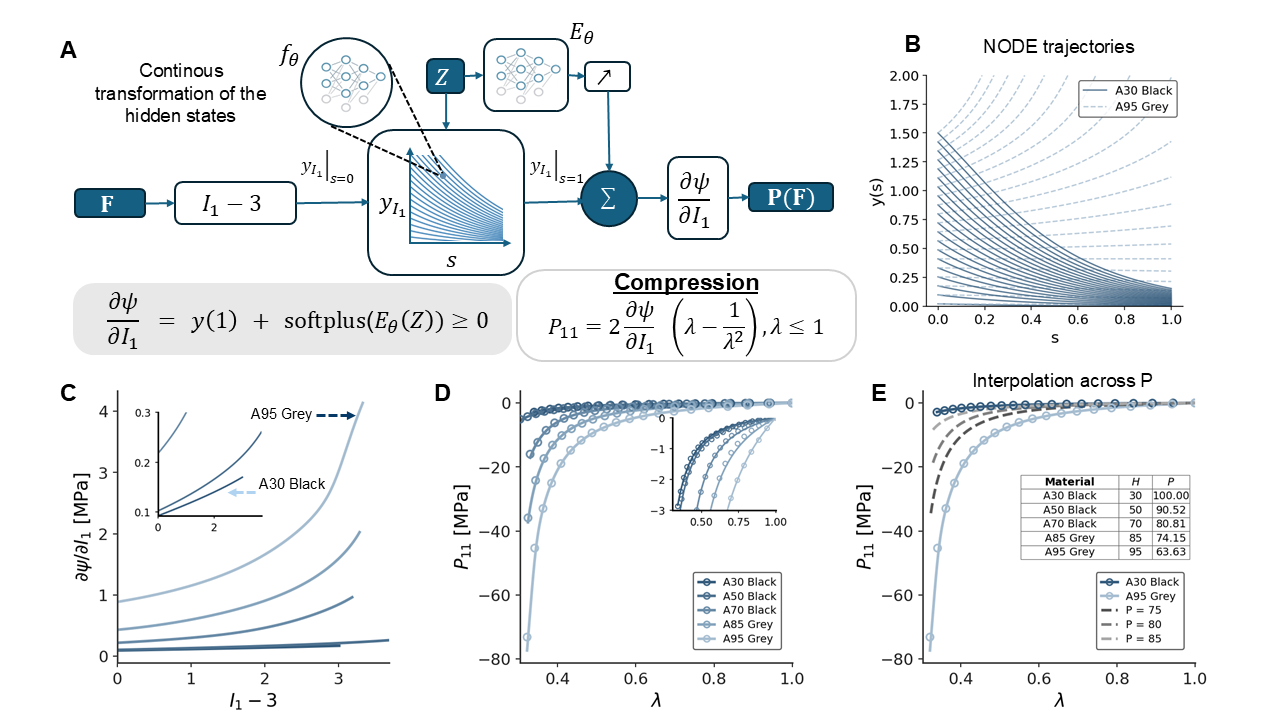}
\caption{\textbf{Multi-material NODE formulation for learning the hyperelastic constitutive response in terms of the invariant} \(\mathbf{I}_{\mathbf{1}}\)\textbf{.} \textbf{(A)} Schematic of the proposed architecture: the deformation gradient \(\mathbf{F}\)is reduced to the invariant measure \(I_{1} - 3\), which is continuously transformed by a NODE conditioned on the material descriptor vector \(Z\); the terminal hidden state is combined with a non-negative offset \(E_{\theta}(Z)\ \)to define \(\partial\psi/\partial I_{1}\), and thus specify the stress \(\bm{P}(\mathbf{F})\). \textbf{(B)} Representative NODE trajectories \(y(s)\) over the artificial integration coordinate \(s \in \lbrack 0,1\rbrack\), showing two admissible transformation regimes of the hidden state between the initial value \(y_{0} = I_{1} - 3\) and the terminal value \(y_{1}\) mappings for different material conditions \(Z\). \textbf{(C)} Resulting learned constitutive scalar \(\partial\psi/\partial I_{1}\) as a function of \(I_{1} - 3\), illustrating the family of nonlinear responses generated by the model. Here, \(y\) is dimensionless, and \(\partial\psi/\partial I_{1}\)is scaled by a fixed reference stress \(\sigma_{ref} = 1\text{\,}MPa\); normalized notation is omitted for clarity. \textbf{(D)} Uniaxial stress prediction \(P_{11}\) versus stretch \(\lambda\), comparing model response against experimental data for multiple digital materials. \textbf{(E)} Interpolation across the material parameter \(P\) (\% Agilus Black), showing predicted \(P_{11}\)--\(\lambda\) responses.}
\label{fig:3}
\end{figure}

NODE allows the exploration of interpolations/extrapolation between different \(P\) values (see \Figref[E]{3}). The \(P\)-conditioned NODE learns a smooth constitutive manifold mapping \((I_{1},P)\) to \(\partial\psi/\partial I_{1}\), Composition acts through two pathways, it modifies the offset \(E_{\theta}(Z)\), which controls the initial stiffness, and the vector field \(f_{\theta}(y,Z)\), which governs the nonlinear evolution of the response. For each fixed composition, the architecture enforces positivity and convexity of the strain-energy function, ensuring mechanical admissibility. The higher Agilus content is the A30 material, \(P = 100\), and is the softest in the database, with \(d\psi/dI_{1}\) on the order of 0.1 MPa. The lower Agilus content, A95 with \(P = 63.63\), reaches \(d\psi/dI_{1} > 4\) MPa at large deformations. The different materials seem to lie along a multi-material response surface parameterized by \(P\), which is what is learned by the NODE framework. Along this surface, the learned NODE map produces smooth and monotonic intermediate \(P_{11}\)-\(\lambda\) responses for unseen \(P\) values. The convexity of \(\psi\) for any \(P\) is part of the architecture design, but the smooth interpolation across composition is not enforced through explicit monotonicity regularization, smoothness penalties, or \emph{post hoc} corrections, indicating that the model has captured a coherent relationship between composition and mechanical response from the data.

\begin{figure}[htbp]
\centering
\includegraphics[width=\linewidth,keepaspectratio]{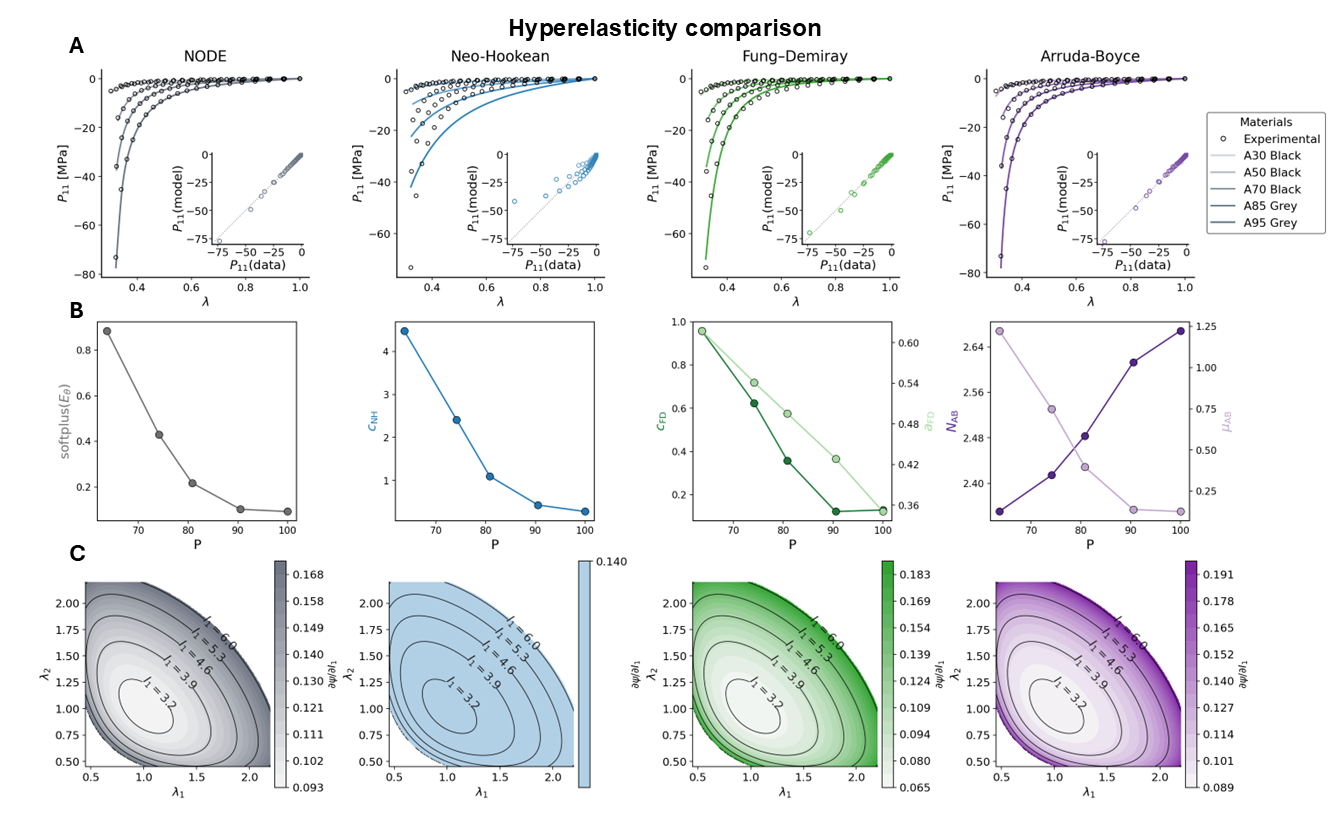}
\caption{\textbf{(A)} Experimental and predicted \(P_{11}\)-\(\lambda\) curves for the NODE, Neo-Hookean, Fung, and Arruda--Boyce models (left to right). Open circles denote experimental data and solid lines denote model predictions for the different materials. \emph{Insets:} parity plots of \(P_{11}^{model}\)versus \(P_{11}^{data}\); closer alignment with the diagonal indicates better agreement. \textbf{(B)} Learned constitutive parameters (see \Tabref{1}) as functions of the material parameter \(P\) (Agilus Black content, \% rubbery material), for the NODE, \(softplus\left( E_{\theta} \right)\) has stress units (MPa) and represents the reference-state value \({\partial\psi/\partial\left. \ I_{1} \right.\mid}_{I_{1} = 3}\), therefore, the small-strain shear modulus is \(G_{NODE} = 2\text{\,}softplus\left( E_{\theta} \right)\). \textbf{(C)} Maps of \(\partial\psi/\partial I_{1}\)on the incompressible \((\lambda_{1},\lambda_{2}\)) plane for a selected material (A30 black). Black curves denote constant-\(I_{1}\) contours.}
\label{fig:4}
\end{figure}

As described in the previous sections, an alternative data-driven strategy is to learn directly the parameters of closed-form constitutive models using MLPs. This comparison between NODE and the closed-form hyperelastic models are shown in \Figref[A]{4}. Neo-Hookean captures the overall monotonic trend of stiffness with composition but cannot reproduce the observed strain stiffening. Fung-Demiray substantially improves the shape of the curves by introducing an exponential function. Arruda-Boyce also improves agreement. The NODE gives the closest overall agreement across materials, and the parity insets show the tightest clustering around the identity line. In \Figref[B]{4}, the parameters conditioned by the latent vector \(Z\) are shown, revealing systematic variations with the material composition parameter \(P\).

To further interpret the constitutive structure beyond the uniaxial curves, we evaluated each equilibrium model on the incompressible principal-stretch plane (\(\lambda_{1},\lambda_{2}\)) where \(\lambda_{3} = (\lambda_{1}\lambda_{2})^{- 1}\) (see \Figref[C]{4}). Since all four models depend only on \(I_{1}(\lambda_{1},\lambda_{2})\), the \(\partial\psi/\partial I_{1}\) isolines coincide with the isolines of \(I_{1}\). The resulting contours illustrate the magnitude and growth of \(\partial\psi/\partial I_{1}\) across a larger set of deformations beyond the uniaxial compression data. In this representation, neo-Hookean remains essentially constant, whereas Fung, Arruda--Boyce, and NODE show a monotonic increase in \(\partial\psi/\partial I_{1}\) with increasing \(I_{1}\), a requirement for the convexity of the energy. While this monotonicity is clear in the closed-form model expressions, it should be noted that the NODE architecture also imposes this condition over the entire principal-stretch plane by the architecture design itself and not by adding terms to the loss function. Hence, even though the NODE model is trained on uniaxial compression data alone, it generalizes well to other deformation modes.

We performed a leave-one-material-out evaluation in which each material was excluded from training and subsequently predicted using the model calibrated on the remaining materials. This train-test split highlights the difference between fitting accuracy and model transfer to unseen materials. As shown in \Figref[A]{5}, NODE generalizes best, whereas Fung--Demiray and unregularized Arruda--Boyce fail more noticeably when extrapolating toward the softer A30 Black material. Introducing a penalty on the normalized chain stretch, \(\chi_{chain}\), substantially improves Arruda-Boyce, leading to the lowest held-out NRMSE across the closed-form materials, and is on par with the NODE framework, with errors averaging ten percent in the validation set. Nonetheless, the summary in \Figref[B]{5} shows that neo-Hookean gives the best individual prediction for A50 Black. This reflects a bias-variance tradeoff in material model interpolation: neo-Hookean is less expressive and cannot capture the strain-stiffening in the training or extrapolation cases, but its limited flexibility makes it more stable under extrapolation. In contrast, more flexible strain-hardening models can exploit poorly constrained parameter regimes during training, leading to high-sensitivity to some parameters and can result in poor prediction when transferred to unseen compositions. \Figref[C]{5} shows the \(P_{11} - \lambda\) plots in the evaluation set across the compositions, from A30 to A95, and highlights the best performer for each case. All predictions group in a small range for A70 to A95 except for neo-Hookean, which stands out as the model without the strain-stiffening response. For A50 and for A30, which is the softest material, there is a more noticeable spread in the predictions, highlighting the difficulty in extrapolation. The NODE model provides the closest predictions to the data in extrapolation.

\begin{figure}[htbp]
\centering
\includegraphics[width=\linewidth,keepaspectratio]{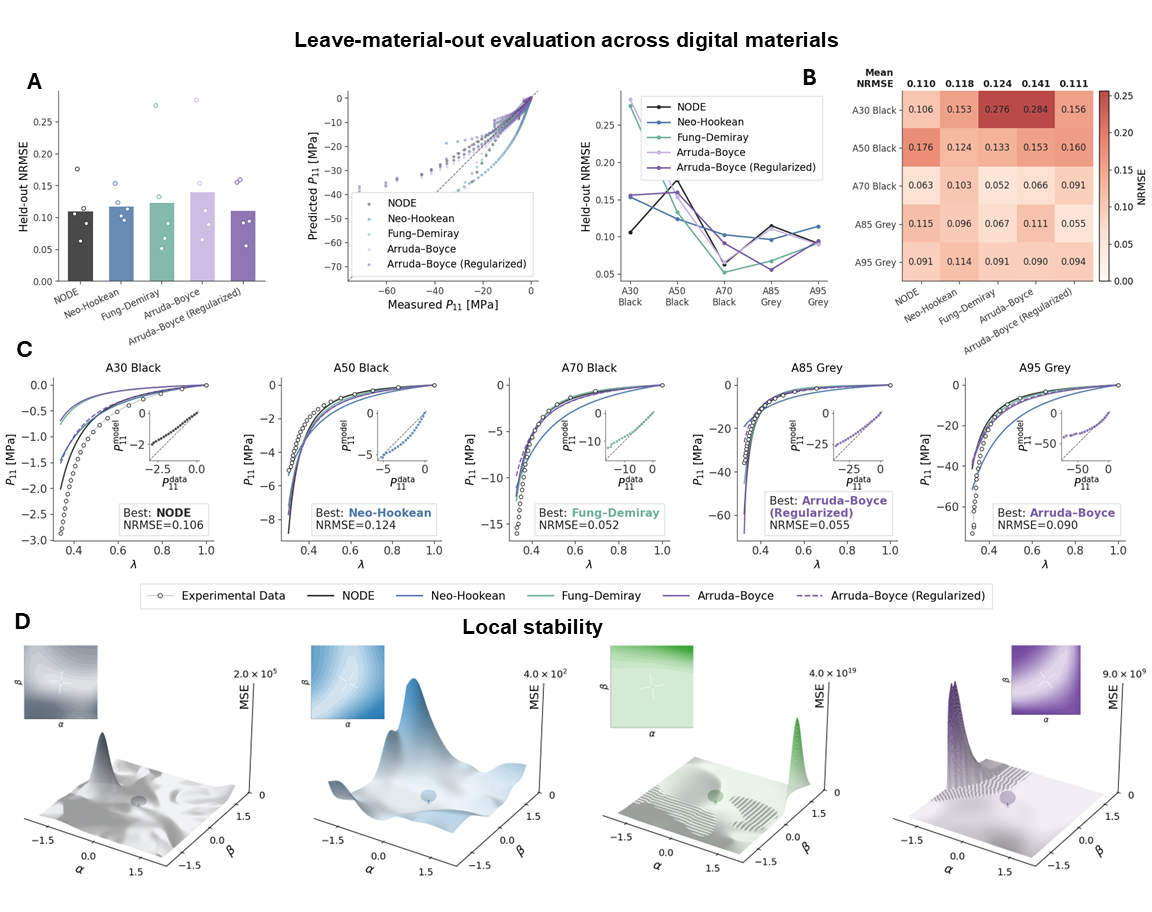}
\caption{\textbf{Leave-material-out evaluation across digital materials. (A)} Held-out NRMSE across models and materials. \textbf{(B)} NODE gives the lowest mean NRMSE, while Arruda--Boyce (with a regularization on the chain stretch) leads to comparable performance and markedly improves over its unregularized training and over the other closed form models. \textbf{(C)} Held-out stress--stretch predictions. The best model depends on the excluded material, showing that model transfer performance is governed not only by model expressivity but also by extrapolation stability and sensitivity to constrained or non-regularized parameter regimes. \textbf{(D)} Local stability. Two-dimensional loss landscapes around the trained solutions (NODE in gray, neo-Hookean in blue, and Fung--Demiray in green, unregularized Arruda--Boyce in purple), obtained by perturbing the parameters along two random directions in parameter space. The vertical axis shows the full-batch mean squared error (MSE). Flatter basins indicate lower local sensitivity, whereas steep peaks and narrow valleys near the optimized parameters centered at \((0,0)\) indicate that the loss optimization is sensitive to perturbations within a non-convex landscape.}
\label{fig:5}
\end{figure}

To determine whether these differences in performance are accompanied by differences in local parameter sensitivity, we evaluated a two-dimensional loss landscape around the trained parameter vector \(\theta^{\star}\) for the neo-Hookean, Fung--Demiray, Arruda--Boyce, and NODE models (see \Figref[D]{5}). The point \(\left( \alpha,\beta) = (0,0 \right)\) corresponds to the trained parameter vector \(\theta^{\star}\) (see \textbf{Appendix E} for details). In these loss landscapes, flatter regions around the optimized value \((0,0)\) indicate lower local sensitivity, whereas narrow valleys and steep peaks indicate stronger dependence on parameter perturbations. The Fung--Demiray and Arruda--Boyce models show relatively flat regions close to the trained parameters, indicating good local stability. However, the MSE increases to much larger values away from this local region compared with NODE and neo-Hookean, indicating greater sensitivity to larger parameter perturbations.

These results provide complementary evidence that physics-based regularization can improve interpolation and extrapolation by restricting the admissible solution space. For instance, the loss landscape for the Arruda-Boyce model without penalty regularization results in some local minima and maxima around the optimized value. The regularized Arruda-Boyce, on the other hand, shows a flat loss around the optimized value, which helps explain is better performance in the evaluation test. Namely, a loss with low sensitivity suggests that the prediction of Arruda-Boyce parameters changes slowly with composition. However, any regularization must be done with care, as constraining parameters without a clear physical interpretation can introduce arbitrary bias rather than physical consistency \citep{Geirhos2020} \citep{Krishnapriyan2021}. Interestingly, the loss landscape for the NODE model is not flat, indicating some sensitivity with respect to the neural network parameters. Yet, this framework extrapolates well to materials not seen during training. A possible reason is that the high-dimensional nature of the NODE parameterization may produce a complex loss landscape while still allowing for a smooth and gradual change in the energy with respect to the network inputs, in particular the composition \(P\).

\subsection*{7.2. Multi-material nonequilibrium response}

To train the viscoelastic response, the parameters of the equilibrium branch \(A\) were fixed from the equilibrium calibration described in the previous section and represented using the Arruda-Boyce model. The nonequilibrium branches were then trained using 20 experimental datasets corresponding to four strain rates across five materials.

\subsubsection*{7.2.1 Comparison of closed-form and NODE energies with a learned evolution law}

In this subsection, we compare two definitions of the branch energy coefficient\(\ \alpha^{(r)}(I_{1,e}^{(r)},Z)\): a closed-form AB-type (Arruda--Boyce strain energy) representation and a NODE representation. The branch evolution law in this subsection is the neural network version, modeled as \({\dot{\gamma}}^{(r)} = \tau_{eff}^{(r)}\text{\,}\phi_{\theta}^{(r)}(\tau_{eff}^{(r)},I_{1,v}^{(r)},Z)\) with \(\phi_{\theta}^{(r)}\) as a positive MLP taking \(\tau_{eff}^{(X)}\) \(I_{1,v}^{(r)}\) and \(Z\) as inputs. This construction guarantees \({\dot{\gamma}}^{(r)} \geq 0\) and therefore preserves the dissipation inequality. With this formulation, a single model captures the broad range of strain rates and material compositions with high accuracy (\Figref[A]{6}). The AB and NODE definitions of \(\alpha^{(r)}\) produce similar global predictions, indicating that both energy representations are compatible with the learned dissipative law. However, at the highest strain rate, the AB model for energy exhibits a sudden nonlinear stiffening along the loading curves of A30 Black, A50 Black, and A70 Black. Branch-level inspection shows that branch B approaches a chain-locking regime in these cases (see \Suppref{19}{Fig.~S3}). This behavior is not unique to the AB model; a similar artifact appears when a Fung-type closed-form representation is used for the elastic energy contribution of the viscoelastic branches (\Suppref{20}{Fig.~S4}). As shown in \Suppref{21}{Fig.~S5}, the training curves converge smoothly and quickly, with no visible spikes in the loss history, even though the objective function contains no explicit regularization and is based only on the curve-level NRMSE. There is comparable convergence across loading rates. This suggests that, in unconstrained optimization of the closed-form models, the optimizer is able to reduce the loss smoothly by driving one branch to have excessive strain hardening. For AB-type models, this issue can be controlled using a physically meaningful penalty on the normalized chain stretch \(\chi_{chain} = \frac{\lambda_{\text{chain}}}{\sqrt{N}}\). This regularization limits unphysical chain locking and removes the loading artifact (\Suppref{21}{Fig.~S6}). Therefore, when using closed-form models with very nonlinear strain hardening (e.g., exponential or inverse Langevin functions), training requires regularization to avoid obtaining a model sitting on the boundary of the parameter space. The NODE energy, possibly due to its flexibility to smoothly interpolate different degrees of strain-stiffening, does not face the same issue.

Even though training was done with a few sets of materials and strain rates, the constitutive model framework allows for interpolation across strain rates (see \Suppref{22}{Fig.~S7}) and intermediate values of the Agilus Black content \(P\) (\Suppref{22}{Fig.~S8}). In the case of the NODE framework, this is achieved because the NODE trajectories are a function of the material feature vector \(Z\) just as in the equilibrium branch. In the closed-form case, the AB parameters are functions of \(Z\) through a MLP, the same strategy used in the equilibrium branch. Regarding dissipation, the learned flow rule, \({\dot{\gamma}}^{(r)} = \tau_{eff}^{(r)}\text{\,}\phi_{\theta}^{(r)}(\tau_{eff}^{(r)},I_{1,v}^{(r)},Z)\), the MLP \(\phi_{\theta}^{(r)}\) in this subsection explicitly takes as an input the features \(Z\) and thus learns the rate-dependence response across the entire material family.

\begin{figure}[htbp]
\centering
\includegraphics[width=\linewidth,keepaspectratio]{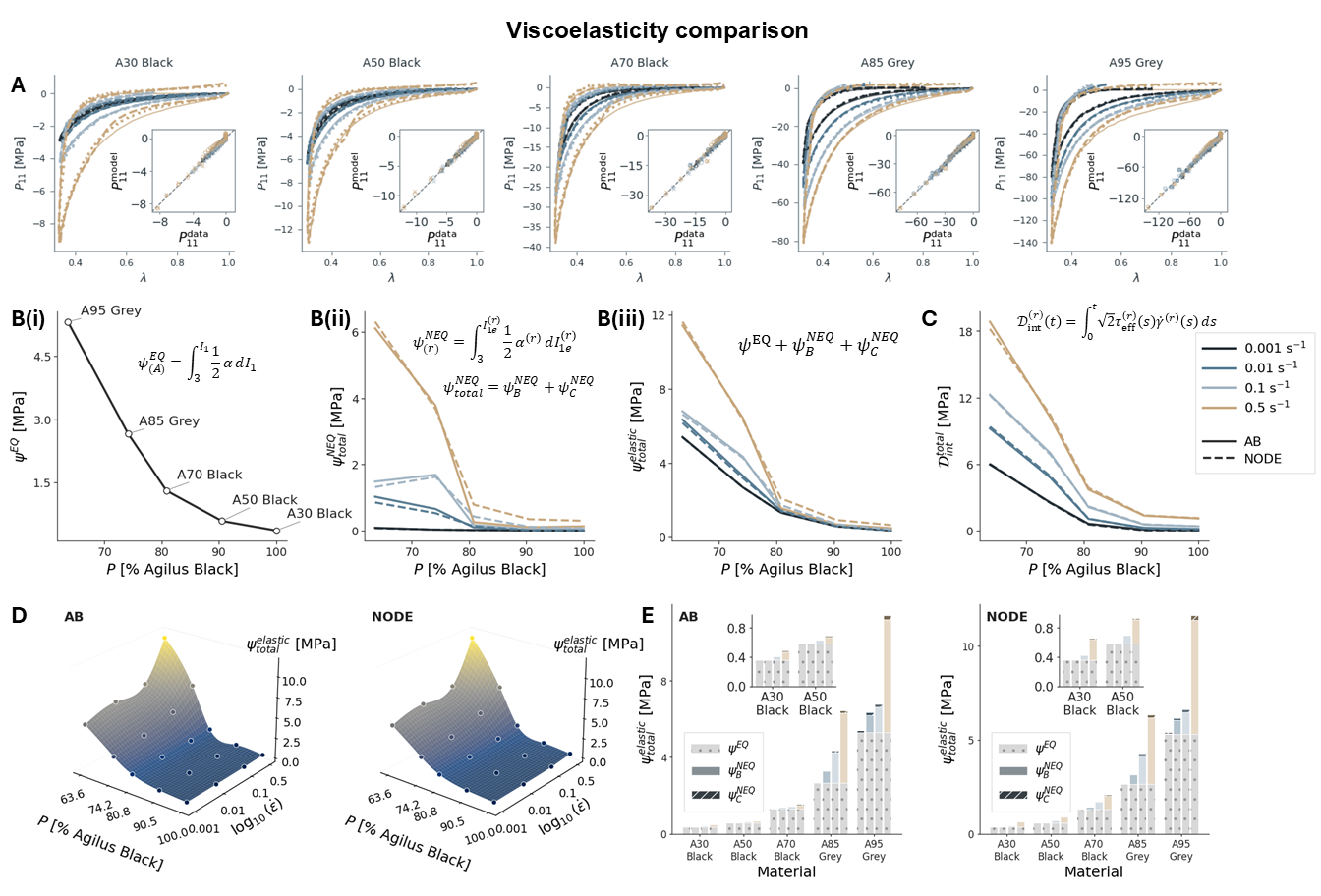}
\caption{\textbf{Multi-material viscoelasticity and energy decomposition. (A)} Experimental and predicted nominal stress--stretch responses for five materials and four strain rates. Solid lines denote experimental data, while dashed and dotted lines denote AB and NODE predictions, respectively. \emph{Insets} show parity plots of predicted versus experimental stress. \textbf{(B-i-ii-ii)} Elastic equilibrium energy (\(\psi^{EQ}\)), total nonequilibrium energy \(\left( \psi_{total}^{NEQ} \right)\) , and total elastic energy \((\psi_{total}^{elastic} = \psi_{total}^{NEQ} + \psi^{EQ})\) at maximum deformation before unloading as a function of strain rate and Agilus content. \textbf{(C)} Cumulative internal dissipation \(\mathcal{D}_{int}^{total} = \mathcal{D}_{int}^{B} + \mathcal{D}_{int}^{C}\), evaluated over the complete loading--unloading cycle as a function of material composition \(P\) and strain rate. \textbf{(D)} Response surfaces of \(\psi_{total}^{elastic}\) over composition (\(P\)) and strain rate at maximum deformation, showing the coupled dependence of energy storage on material composition and loading rate. \textbf{(E)} Branch-wise energy decomposition for AB and NODE models, separating equilibrium and nonequilibrium contributions across materials and strain rates.}
\label{fig:6}
\end{figure}

The calibrated multi-material model enables analysis beyond stress-stretch predictions. In particular, we are interested in how the energy is stored and dissipated in these materials. As discussed above, the equilibrium energy increases nonlinearly as the Agilus Black content decreases (\Figref[B-i]{6}). When considering the non-equilibrium response, we see that the nonequilibrium energy shows an even stronger dependence on composition and strain rate, with high-rate loading producing a substantial increase in the stored nonequilibrium contribution for materials with lower Agilus Black content (\Figref[B-ii]{6}). Adding both contributions, it is further clear that lower Agilus Black content (lower \(P\), e.g. A95 Grey) leads to larger equilibrium and nonequilibrium energy contributions (\Figref[B-iii]{6}). Note that, for consistency, all stored energy quantities are evaluated at the maximum deformation during the loading, which is shared among all tests.

The monotonic increase in stored elastic energy with strain rate is expected since faster loading gives less time for viscoelastic relaxation before reaching maximum deformation. This rate-dependent increase is small for A30 Black but becomes much more pronounced toward A95 Grey, as shown by the larger separation between the strain-rate curves in \Figref[B-ii--iii]{6}. To understand the balance between stored energy and dissipation in the non-equilibrium branches, we also computed the internal dissipation (\Figref[C]{6}). Dissipation follows largely the same trends as the stored elastic energy. Materials with lower Agilus Black content (lower \(P\), e.g., A95 Grey) are more dissipative, with greater dissipation at larger strain rates. In some cases, the dissipated energy even exceeds the total stored elastic energy. Therefore, materials with lower Agilus Black content show both larger stored-energy contributions and greater dissipation than materials with higher Agilus Black content.

The response surfaces in \Figref[D]{6} summarize the coupled dependence on \(P\) and strain rate, enabling the identification of compositions with targeted viscoelastic performance. In the proposed modeling framework, two non-equilibrium branches are considered due to the inability of a single non-equilibrium branch to capture the observed behavior (\Suppref{23}{Fig.~S9}). At the branch level, the learned energy split between the B and C non-equilibrium branches shows that branch B accumulates a large contribution of the total elastic energy and is highly sensitive to strain rate (see \Figref[E]{6}). Branch C, in contrast, has almost no stored energy contribution. Although AB and NODE use different approaches to describe the energy and there is no regularization to encode a particular branch-wise behavior, \Suppref{24}{Fig.~S10} shows that their branch-wise nonequilibrium-energy trajectories follow similar features. These results suggest that this material family has two distinct dissipation mechanisms discovered by our multi-material modeling framework.

\begin{figure}[htbp]
\centering
\includegraphics[width=0.80\linewidth,keepaspectratio]{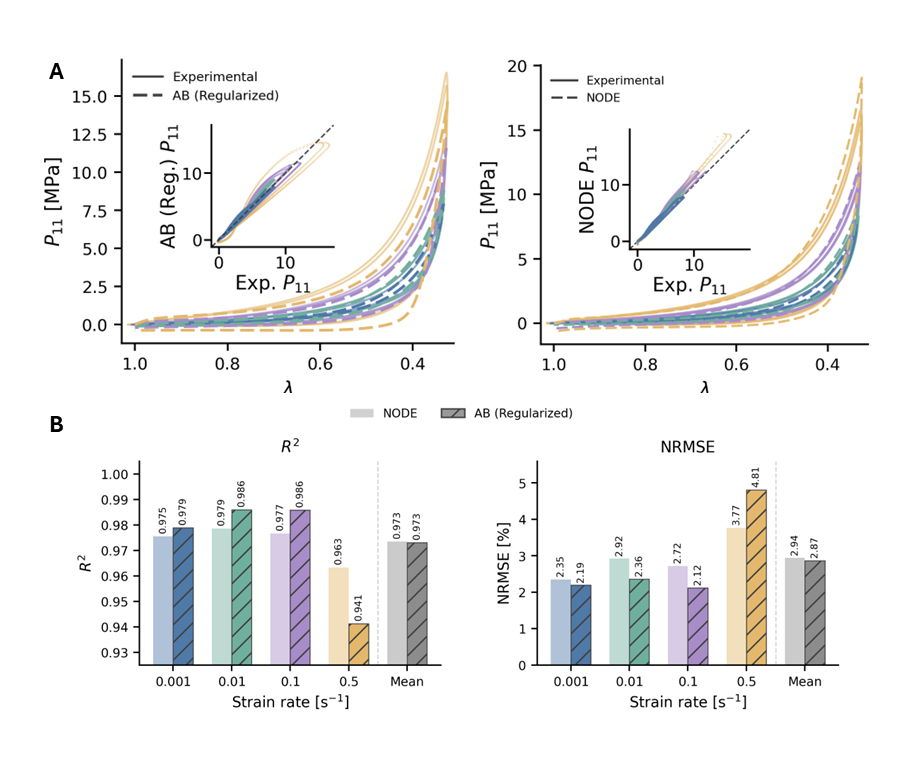}
\caption{Held-out predictions for A60 Black using regularized Arruda--Boyce (AB) and NODE, both with the same positive neural-network flow rule. \textbf{(A)} Experimental and predicted stress--stretch curves. \textbf{(B)} R\textsuperscript{2} and NRMSE by strain rate and mean value.}
\label{fig:7}
\end{figure}

To further assess transfer to an unseen composition (A60 Black), we excluded it from training and evaluated both the regularized Arruda--Boyce (AB) and NODE viscoelastic models on this held-out material, using the same positive neural-network flow rule in both cases (\Figref[A]{7}). Both models achieve comparable mean performance across strain rates. AB performs slightly better at the three lower rates, but its performance degrades at 0.5s\textsuperscript{-1} (\Figref[B]{7}). At this highest rate, AB tends to underestimate the loading response and shows a visible artifact during unloading, associated with the regularization trade-off used to control strain hardening in the elastic energy (see the unloading curve for A70 Black in \Suppref{21}{Fig.~S6B}), whereas NODE slightly overestimates parts of the loading response but gives a smoother unloading curve with a lower error.

\begin{figure}[htbp]
\centering
\includegraphics[width=\linewidth,height=0.72\textheight,keepaspectratio]{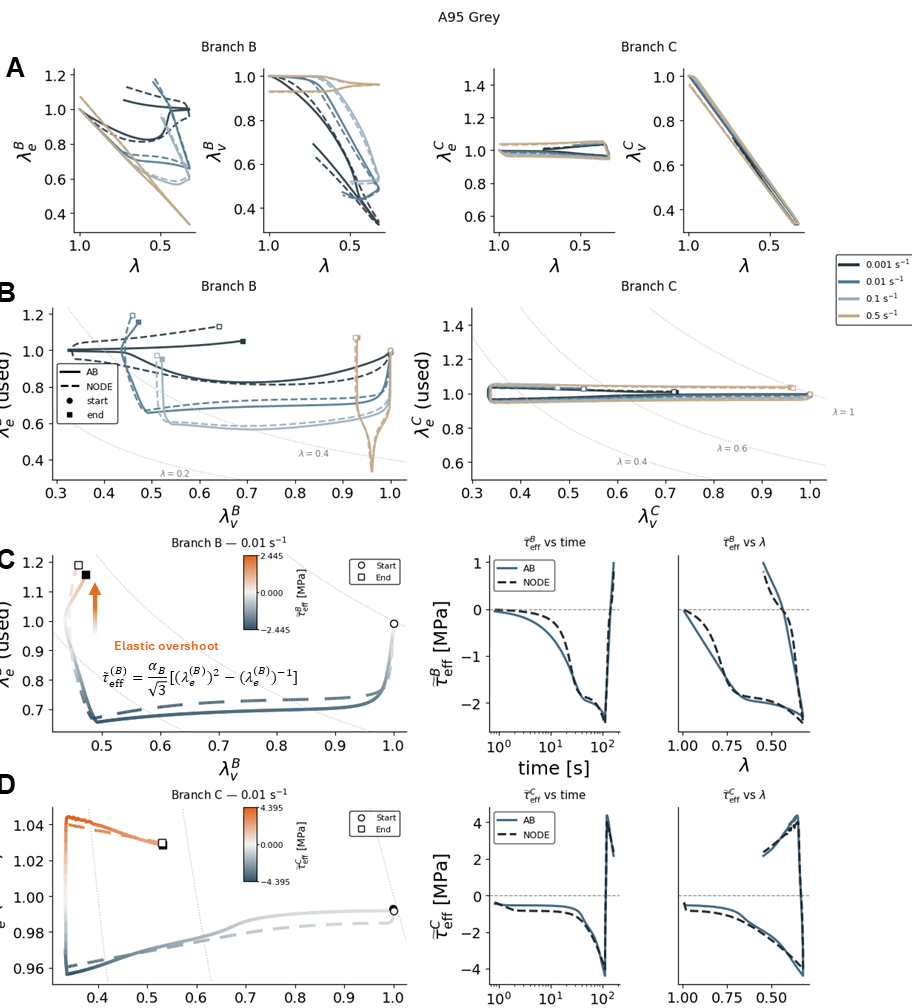}
\caption{\textbf{(A)} Elastic and viscous internal stretches, \(\lambda_{e}^{(r)}\) and \(\lambda_{v}^{(r)}\), plotted against the macroscopic stretch \(\lambda\) for branches \(r = \{ B,C\}\) for A95 Grey, across loading rates. \textbf{(B)} Corresponding internal-state phase portraits in the \((\lambda_{v}^{(r)},\lambda_{e}^{(r)})\) plane; gray dotted curves indicate iso-\(\lambda\) lines. \textbf{(C-D)} Branch B and branch C phase portraits for rate 0.01 s\textsuperscript{-1}, colored by the signed effective stress \({\widetilde{\tau}}_{eff}^{C}\), highlighting the region of elastic overshoot (tension loading). The right panels show the evolution of \({\widetilde{\tau}}_{eff}^{B}\) versus time and versus \(\lambda\). Comparison of the AB and NODE models shows that branches B and C develop tension during unloading even though the overall material is still under compression (\(\lambda < 1)\).}
\label{fig:8}
\end{figure}

Beyond the prediction accuracy, further insight can be obtained by examining the elastic and viscous responses of each branch. We focus on the material with the lowest \(P\) content because it shows the largest accumulation of energy as well as the largest dissipation at higher strain rates. \Figref[A]{8} shows the elastic and viscous responses of both branches for A95 Grey material as a function of the total applied stretch, revealing a clear separation between branches B and C. The internal-variable state phase portrait, (trajectory plot in the (\(\lambda_{v}^{(r)},\lambda_{e}^{(r)}\)) plane) further illustrates elastic versus dissipative behavior (\Figref[B]{8}). Notice that branch C elastic deformation remains close to 1, which can be seen in both visualizations \Figref[A,B]{8}. Consequently, the viscous stretch in this branch, \(\lambda_{v}^{C}\), follows closely the total stretch (\Figref[A]{8}), indicating that branch C accumulates little elastic energy and, instead, it mostly dissipates energy, even at high strain rates.

In contrast, branch B exhibits large history-dependent loops in the \((\lambda_{v}^{B},\lambda_{e}^{B})\) phase portrait, with marked rate sensitivity. This behavior is consistent with branch B storing most of the nonequilibrium energy, as shown already in \Figref[E]{6}. For intermediate rates, branch B shows a rapid accumulation of stored elastic energy at the start of the deformation. Then, dissipation kicks in, trajectories in the (\(\lambda_{v}^{B},\lambda_{e}^{B}\)) space flatten and viscous deformation accumulates. During unloading, the deformation is mostly elastic again. These loops in (\(\lambda_{v}^{B},\lambda_{e}^{B}\)) space in branch B are rate-dependent, with lower loading rates showing more dissipation. During uniaxial compression, the intensity and direction of the elastic response are represented through \({\widetilde{\tau}}_{eff}^{B}\). Notice that for some rates, there is substantial overshoot of \(\lambda_{e}^{B}\) above 1, leading to the branch being in tension, \({\widetilde{\tau}}_{eff}^{B} > 0\) (\Figref[C]{8}). In general, branch B dissipates energy slowly enough to store loading history and can become loaded in tension even though the overall material is still under compression, whereas branch C dissipates most of the energy, i.e. it has a more rapid timescale, but it also shows a stiffer mechanism, reflected in the larger \({\widetilde{\tau}}_{eff}^{C}\) values (\Figref[D]{8} and also \Suppref{19}{Fig.~S3B}). This larger stress in branch C, despite its small elastic deformation, is explained by its much larger elastic stiffness compared with branch B (see \Suppref{19}{Fig.~S3B}).

The complex dissipation response summarized in \hyperref[fig:6]{\textbf{Figs.~\ref*{fig:6}}}, \hyperref[fig:7]{\textbf{\ref*{fig:7}}}, \hyperref[fig:8]{\textbf{\ref*{fig:8}}} is learned with a positive neural network for \(\dot{\gamma}\) (\textbf{Eq. 3.23}), using either AB or NODE models for elastic energy. Regardless of whether the elastic energy is captured with a closed-form or the purely data-driven NODE model, the same viscous flow \(\dot{\gamma}\) is learned. In the next subsection we restrict the dissipation to be semi-closed form.

\subsubsection*{7.2.2 Semi-closed evolution laws: interpretability and identifiability}

To complete the study, we replaced the neural-network evolution law with semi-closed-form expressions. Among the semi-closed form alternatives, the selected forms summarized in \Tabref{4} provide the simplest candidates while six additional variants are reported in \Suppref{2}{Table~S2} of the Supplementary Material, including Option 1 which corresponds to the most general form in \textbf{Eq. (3.19)}. Option 5 shows greater stability during training, whereas Option 2, although less stable, achieves the lowest error after 1000 epochs (see \Suppref{25}{Fig.~S11}).

These evolution laws are termed semi-closed form because they include parameters that are learned as a function of composition as well as functions of the material descriptor \(Z\) and of the internal state \(u^{(r)}\). However, as opposed to the fully data-driven approach, \Tabref{4} shows explicit assumptions regarding the linear or power-law dependence of the flow rule on the non-equilibrium stress. The learned parameters of the models picked in \Tabref{4} are shown in \Figref[A]{9}. Different parameter combinations can produce nearly indistinguishable macroscopic responses across materials and loading rates (see \Figref[B]{9}). This behavior is consistent with observations in the equilibrium branch analysis that constitutive parameters may remain non-identifiable from limited mechanical tests, even when the fitted response is accurate, and that material-model calibration can exhibit parameter covariance and non-unique solutions \citep{Safa2021}. This creates compensatory trade-offs between the parameters such as \(K^{(r)}\) (inverse of viscosity), the stress-sensitivity exponents \(m^{(r)}\) , and the dependence on the internal state \(f(u^{(r)})\) (strain hardening or softening). In some cases, viscous features that were previously concentrated in one branch are redistributed across branches, and the elastic overshoot becomes more pronounced (see \Suppref{26}{Fig.~S12}).

Option 2 forces branch B to a linear relationship between nonequilibrium stress and viscous flow, and \Figref[A]{9} identifies that inverse-viscosity \(K^{(B)}\) scales proportionally to \(P\). The viscosity is modulated by a strain-hardening response \(f_{1}^{(B)}\). Option 2, branch C, allows for a power law, and the identified models show that for higher \(P\) (e.g., A30 Black), the response is almost linear (\(m^{(C)}\sim 1\)), but the exponent rises with decreasing Agilus content. For A95 Grey, \(m^{(C)}\sim 6\). Interestingly, while for the neural network dissipation model there was little effect if the energy was closed-form or NODE, for the semi-form dissipation model, the choice of energy representation has a substantial effect. In Option 2, for example, the viscous hardening between branches B and C is flipped depending on whether NODE or AB models are used for the energy.

Option 5 allows for power law dissipation in both branches, but it restricts the viscous flow hardening to branch C. The identified parameters show that in this case, both branches rely on higher exponent for lower Agilus Black content (\(m^{(B)},m^{(C)}\sim 3\) for A95 material), but reduce to almost linear dependence on the stress for higher \(P\) (\(m^{(B)},m^{(C)}\sim 1\) for A30 material). Interestingly, to capture the data, Option 5 identifies a non-monotonic viscous flow-hardening \(f_{2}^{(C)}(\lambda_{chain,v}^{C})\). Its initial increase suppresses viscous flow and therefore represents viscous hardening at small viscous stretch, whereas its subsequent decrease facilitates flow and indicates a transition toward viscous softening. Because Option 5 does not include a corresponding state-dependent function in branch B, \(f_{2}^{(C)}\) alone absorbs all deformation dependence.

The key point is therefore that the choice of functional form imposes strong assumptions, yet it can still be general enough that it compromises identifiability, all while accurately capturing the observed stress-stretch behavior. Additional constraints could be introduced to improve identifiability and physical interpretation, such as enforcing monotonicity of \(f(u^{(r)})\), bounding the exponents \(m^{(r)}\), or restricting the relative activity of each branch. However, these constraints must be imposed with care, since they can introduce model bias when the assumptions are stronger than what is supported by the data. For example, if monotonicity of the state function is required, Option 5 can be discarded. In this scenario, Option 2 implies the need for two dissipative branches both with strain hardening, decreasing viscosity (inverse of \(K^{(B)},K^{(C)}\)) with higher Agilus Black content, and power-law behavior for branch C with greater exponent for lower Agilus Black content.

\hypertarget{tab:4}{\textbf{Table 4.}} Proposed semi-closed evolution laws for branches B and C.

\begin{longtable}[]{@{}
  >{\raggedright\arraybackslash}p{(\columnwidth - 4\tabcolsep) * \real{0.2349}}
  >{\raggedright\arraybackslash}p{(\columnwidth - 4\tabcolsep) * \real{0.3614}}
  >{\raggedright\arraybackslash}p{(\columnwidth - 4\tabcolsep) * \real{0.4037}}@{}}
\toprule\noalign{}
\begin{minipage}[b]{\linewidth}\raggedright
\end{minipage} & \begin{minipage}[b]{\linewidth}\raggedright
\textbf{Branch B}
\end{minipage} & \begin{minipage}[b]{\linewidth}\raggedright
\textbf{Branch C}
\end{minipage} \\
\midrule\noalign{}
\endhead
\bottomrule\noalign{}
\endlastfoot
\textbf{Option 2} & \({\dot{\gamma}}^{(B)} = K^{(B)}(Z)\text{\,} \times \frac{\tau_{eff}^{(B)}}{f_{1}^{(B)}\left( u^{(B)} \right)}\) & \({\dot{\gamma}}^{(C)} = K^{(C)}(Z)\text{\,} \times \left( \frac{\tau_{eff}^{(C)}}{f_{2}^{(C)}\left( u^{(C)} \right)} \right)^{m^{(C)}(Z)}\) \\
\textbf{Option 5} & \({\dot{\gamma}}^{(B)} = K^{(B)}(Z)\text{\,} \times \left( \frac{\tau_{eff}^{(B)}}{\tau_{ref}} \right)^{m^{(B)}(Z)}\) & \({\dot{\gamma}}^{(C)} = K^{(C)}(Z)\text{\,} \times \left( \frac{\tau_{eff}^{(C)}}{f_{2}^{(C)}\left( u^{(C)} \right)} \right)^{m^{(C)}(Z)}\) \\
\end{longtable}

\begin{figure}[htbp]
\centering
\includegraphics[width=\linewidth,keepaspectratio]{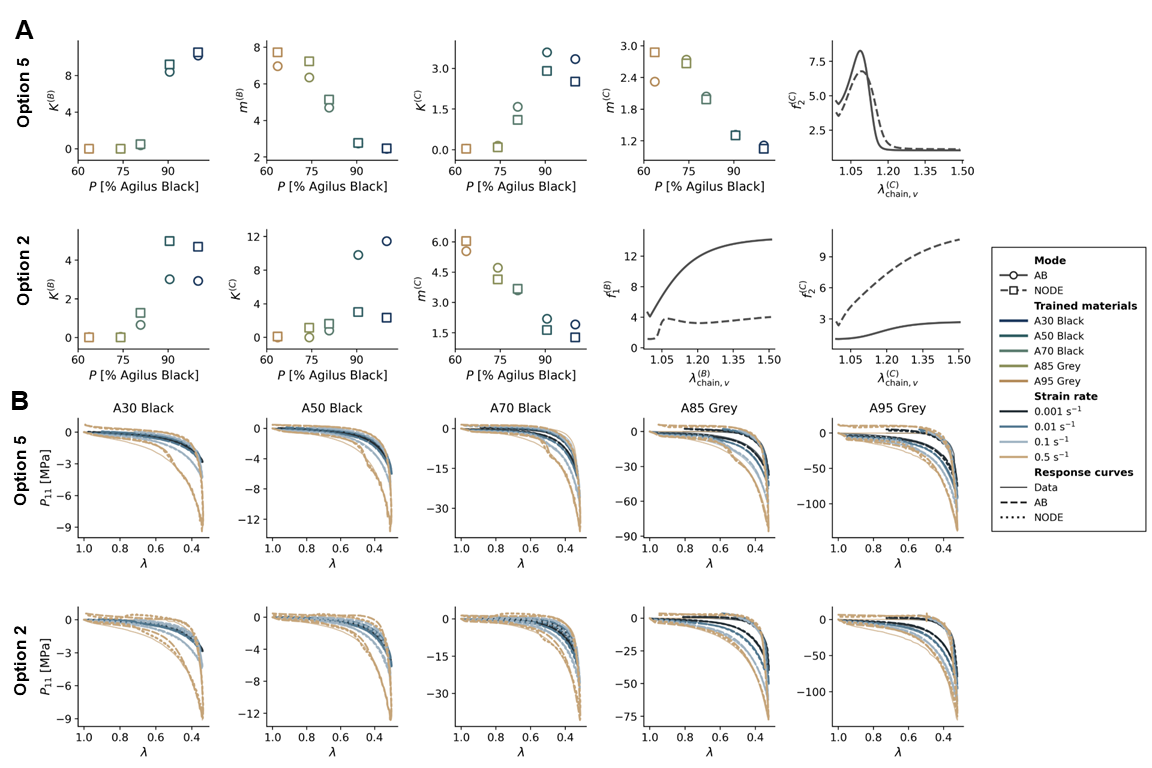}
\caption{\textbf{(A)} Learned semi-closed parameters \(K^{(B)}\), \(m^{(B)}\), \(K^{(C)}\), and \(m^{(C)}\) across the trained materials, together with \(f_{1}^{(B)}\) and/or \(f_{2}^{(C)}\), for the candidate evolution laws described in \Tabref{4}. \textbf{(B)} AB and NODE stress predictions with experimental \(P_{11}\)--\(\lambda\) responses across materials and strain rates.}
\label{fig:9}
\end{figure}

\section*{8. Conclusion and future work}

This work presents a data-driven multi-material framework for the nonlinear, strain-rate-dependent response of digital materials, in which neural networks either learn composition-dependent parameters of closed-form strain energies and flow rules, or represent the constitutive laws directly through NODEs and positive neural networks conditioned on material features. At equilibrium, NODE and a regularized Arruda-Boyce model perform best, capturing pronounced strain stiffening across orders of magnitude in stress-stretch responses. The comparison across closed-form and data-driven models exposes a real trade-off in multi-material constitutive modeling. A parsimonious, interpretable model is only an advantage if it captures the right physics. The neo-Hookean model stays stable under interpolation and extrapolation but cannot reproduce the observed strain stiffening, while Arruda-Boyce has the right micromechanical structure. Nevertheless, using Arruda-Boyce without regularization, unconstrained optimization can still push its parameters toward the edge of a physically meaningful range to compensate for small model deficiencies. Regularization terms can correct the multi-material extension guided by the underlying mechanics, but this requires expert modeling judgment. NODE, by contrast, gives up interpretability and a compact parameter count in exchange for flexibility, leading to accurate and robust interpolation and extrapolation by learning a smooth constitutive manifold over a large space of material responses. It should be noted that the NODE model is extremely flexible, but it is not an arbitrary neural architecture. Like other recent PANNs, it has physical and mathematical requirements such as objectivity and polyconvexity built in \citep{Klein2022b,Linka2023,Abdolazizi2024}.

The same trade-offs between interpretability and robustness appear in the nonequilibrium branches. The Agilus-Vero digital materials show stress and dissipation that grow substantially with strain rate. A strain energy based on the Arruda-Boyce model combined with power-law dissipation has the right physics but again requires careful regularization to avoid the chain-locking artifacts produced by unconstrained optimization. In contrast, the NODE energy paired with a positive-MLP flow rule is flexible enough to learn a smooth manifold across compositions and rates without further regularization. Even in this fully data-driven configuration, the model is not completely unconstrained. There are key assumptions that underpin the entire framework, most crucially the multiplicative kinematics and the flow rule directed along the normalized nonequilibrium deviatoric stress \citep{Bergstrm1998,Bergstrm2000,Bergstrm2001}. NODEs and MLPs describe only a handful of scalar constitutive relationships within this thermodynamically consistent structure. Within that structure, the flexibility of the positive-MLP flow rule consistently separates the nonequilibrium response into two branches with distinct dissipative roles. One branch evolves rapidly, accommodating most of its deformation through viscous flow while storing little elastic energy. The other branch evolves more slowly, stores most of the nonequilibrium elastic energy, and retains directional loading history. This separation was not imposed \emph{a priori}, and it holds whether the branch elasticity is represented with closed-form or NODE representations, which suggests it is a real constitutive feature of these materials rather than a fitting artifact. While we do not investigate the micromechanical origins of this behavior, a possible explanation is that the slow dissipation in one of the branches could be long-timescale reptational relaxation of the entangled network, dominant at low rates \citep{Likhtman2002}. The fast relaxation in the other branch can be expected from local, deformation-induced transient disentanglement or other short-time chain rearrangements \citep{Zhou2018}. A companion paper \citep{Shen2026} investigates this hypothesis directly, combining dynamic mechanical analysis with a broader compositional range extending into the glassy regime to develop a bottom-up, micromechanically motivated framework.

Overall, our results support using PANNs as tools for constitutive discovery for finite viscoelasticity of polymers and other soft material families \citep{Jones2026,Kalina2026}. Natural next steps include extending the framework to multiaxial loading data as well as damage and fracture of digital materials.

\section*{Authorship contribution}

JG-A: Conceptualization, Methodology, Investigation, Formal analysis, Data curation, Software, Validation, Visualization, Writing -- original draft, Writing -- review \& editing.

BS: Formal analysis, Data curation, Writing -- review \& editing.

MKR: Methodology, Funding acquisition, Resources, Writing -- review \& editing.

MCB: Methodology, Supervision, Resources, Writing -- review \& editing.

ABT: Conceptualization, Methodology, Supervision, Funding acquisition, Resources, Writing -- review \& editing.

\section*{Declarations}

The authors have no conflicts of interest to declare.

\section*{Acknowledgments}

ABT acknowledges support of ARO under awards W911NF-24-1-0244, W911NF-26-1-A010. ABT and MKR also acknowledge support from ONR under award N00014-251-2237.

\FloatBarrier

\end{document}